%% file: arxiv.tex
\documentclass[10pt,twocolumn,letterpaper]{article}

\usepackage[pagenumbers]{cvpr}      %

\usepackage{caption}  %
\usepackage{xspace}
\usepackage{fvextra}  %

\newcommand{\method}{DemoFlow\xspace}

\input{numbers}  %

\definecolor{linkblue}{rgb}{0.21,0.49,0.74}
\usepackage[pagebackref,breaklinks,colorlinks,allcolors=linkblue]{hyperref}

\title{DemoFlow: From One Demonstration to Unseen Interactions via 3D Flow}

\author{
Sangjun Noh$^{1*}$ \quad Dongwoo Nam$^{2*}$ \quad Seunghyeok Hyun$^{2}$ \quad Kangmin Kim$^{2}$ \quad Kyoobin Lee$^{2\dagger}$\\
$^{1}$Korea Institute of Science and Technology (KIST)\\
$^{2}$Gwangju Institute of Science and Technology (GIST)\\
{\tt\small sangjun7@kist.re.kr, \{namdongwoo, kyoobinlee\}@gist.ac.kr}
}

\begin{document}

\twocolumn[{%
\renewcommand\twocolumn[1][]{#1}%
\maketitle
\begin{center}
\includegraphics[width=\linewidth]{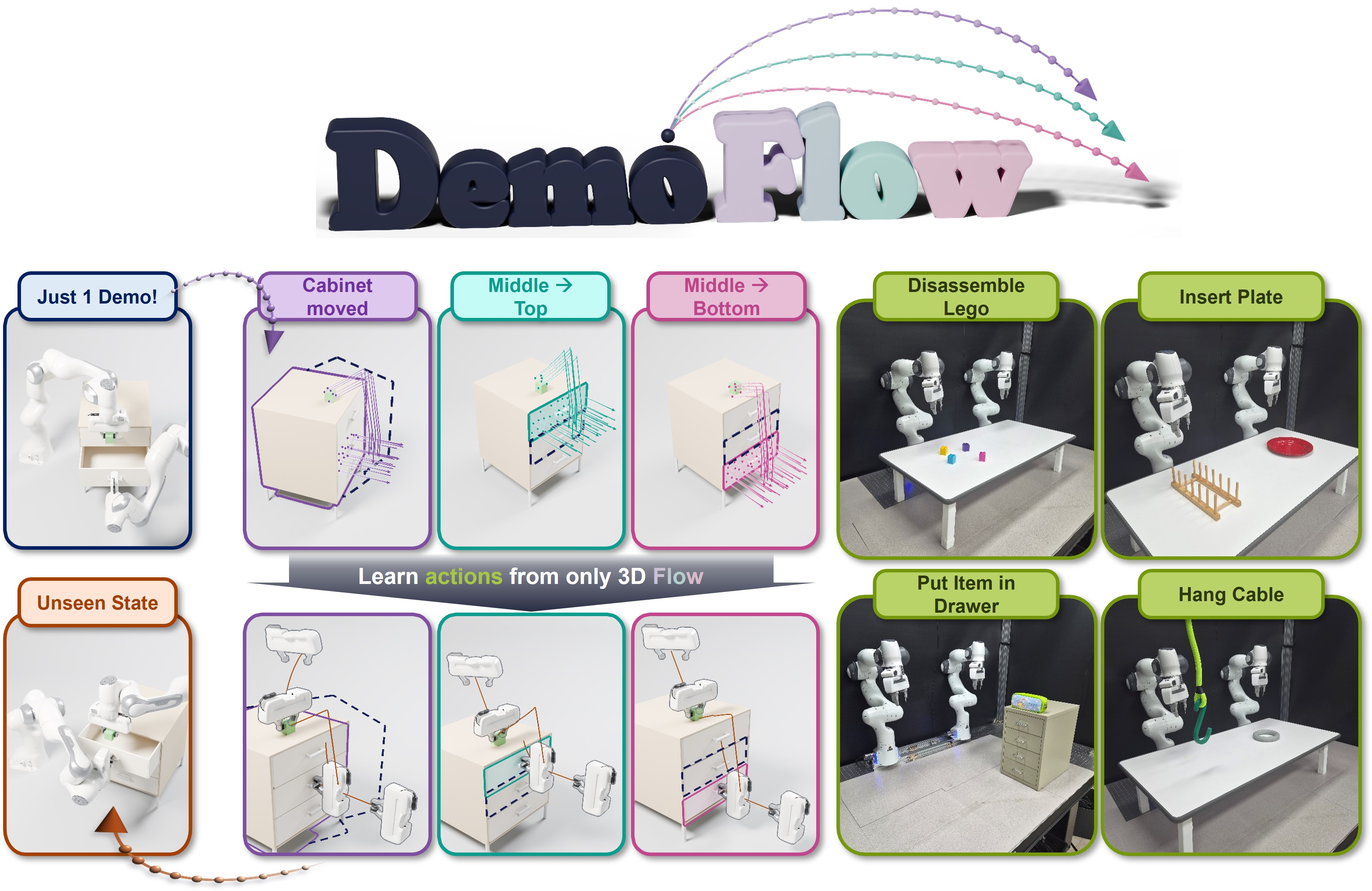}
\captionsetup{hypcap=false}
\captionof{figure}{\textbf{From one demonstration to unseen interactions.}
\method edits the 3D flow and the action of one demonstration into new demonstrations, such as a moved cabinet or another drawer.
A policy that predicts its actions only from the flow then performs them in unseen scenes.
Right: the four real-robot tasks.}
\label{fig:teaser}
\end{center}
}]
{\let\thefootnote\relax\footnotetext{$^*$Equal contribution. \quad $^\dagger$Corresponding author.}}

\begin{abstract}
Imitation learning trains a robot policy directly from human demonstrations and has become a common way to teach manipulation skills.
However, a learned policy rarely succeeds beyond the scenes that its demonstrations cover, so every new object placement, instructed target, or object state calls for more demonstrations, which are costly to collect. %
Existing one-shot methods edit a single demonstration into new ones by transforming segmented objects with rigid poses, which suits new placements but does not change which part moves or the shape it starts in.
Meanwhile, 3D flow, which records where each point in the scene moves over the next steps, has been used to transfer motion across embodiments or to strengthen the observation features of a policy.
Our key insight is that 3D flow can be the medium of one-shot imitation itself.
Since a flow records how every point moves without grouping the points into objects, one point-wise edit of the points, their flow, and the robot hands creates a new demonstration, such as a moved scene, another drawer of the same cabinet, or a new rope shape.
We present \textbf{\method}, which edits the 3D flow of a single demonstration and trains a policy that predicts its actions only from the flow.
On 12 bimanual RLBench2 tasks, one edited demonstration raises the average success rate from \OneShotBaseMean\% to \OneShotEditMean\%, and its average is close to that of 25 unedited demonstrations. %
On a real bimanual robot, one edited demonstration outperforms 25 unedited demonstrations on all four tasks, with an average success rate of \RealEditOneMean\% against \RealNoEditTwentyFiveMean\%.
Project page: \url{https://gistailab.github.io/demoflow/}
\end{abstract}

\section{Introduction}
\label{sec:intro}

Imitation learning enables robots to acquire manipulation skills from human demonstrations~\cite{chi2024diffusion,zhao2023learning,ze2024dp3}.
However, a policy often fails in scenes that its demonstrations do not cover, and collecting demonstrations of many object positions, targets, and states is costly for every new task.
To reduce this cost, prior work generates demonstrations in simulation, either with simulation tasks and code written by language models~\cite{wang2024gensim,hua2024gensim2,ha2023scaling,mu2025robotwin,chen2025robotwin} or by replaying transformed demonstrations~\cite{mandlekar2023mimicgen,garrett2025skillmimicgen,jiang2025dexmimicgen}, which needs scripted primitives or several annotated source demonstrations.
One-shot editors go further and edit a single demonstration directly.
DemoGen and its successors~\cite{xue2025demogen,xu2026r2rgen,zhou2025you} segment the objects in a point cloud and transform the objects and the hand trajectory with a rigid pose, which works well for new placements.
However, many changes are not a new placement.
Opening another drawer of the same cabinet needs the motion of a part that the demonstration never moved, and a rope that starts bent is not a rigid change at all.

A 3D flow gives the future 3D positions of points sampled in the scene.
Prior policies predict the flow and follow it, often to transfer motion across embodiments~\cite{eisner2022flowbot3d,yuan2025general,xu2025flow,zhi20253dflowaction}, or predict it with the action as a signal that strengthens the observation features~\cite{yang2025gripper,chen2026rotridiff,noh20253d,wang2026lamp}.
None of them edits the flow to create training data.
We observe that a flow is also a natural space for editing.
Since a flow does not group the points into objects, one point-wise operation on the points, their flow, and the hands can change not only where an object is but also its state.
The same operation moves the whole scene, transplants the opening motion of one drawer to another drawer, and bends the start shape of a rope (\cref{fig:teaser,fig:edit_all}); the rigid pose edit of prior editors is the special case that applies one transform to all points. %

Based on this observation, we propose \method, which uses 3D flow as the medium of one-shot imitation.
\method applies four edit skills to one demonstration, a pose, a color, an articulation, and a deformation edit.
The edits also need knowledge that a flow does not record, such as which item is the red one or which scene motions keep the task feasible.
A vision-language model (VLM) brings this knowledge from the first frame, naming the targets of the edits and writing a short program that bounds how far the pose edit may move the scene, while the motion comes from the demonstrated flow. %
To learn these diverse edits, the policy uses 3D flow as its intermediate representation.
A 3D flow generator predicts the flow of the scene from the observation, and an action generator predicts the action from only this flow and the robot state.
The flow is thus both what we edit and the only scene input of the action generator. %
Flow-conditioned policies such as PPI~\cite{yang2025gripper} instead read the full observation features, and their action can ignore a changed flow (\cref{sec:exp:follow}). %

We evaluate \method on 12 bimanual tasks of RLBench2~\cite{grotz2024peract2} and on four real-robot tasks.
In simulation, one edited demonstration raises the average success rate from \OneShotBaseMean\% to \OneShotEditMean\%, and on their tasks the articulation and color edits add \LooDropDrawerAvg{} and \LooDropHandHardAvg{} percentage points over the pose edit alone (\cref{sec:exp:beyond}). %
Our contributions are as follows:
\begin{itemize}
\item We propose \method, which creates training demonstrations by editing 3D flow and extends one-shot editing beyond rigid placement to articulated and deformable objects. %
\item We design a two-stage policy whose action generator reads only the flow and the robot state, so that it learns each edit through the flow; in our interventions, its action follows a changed flow more than that of PPI. %
\item We show that one edited demonstration reaches an average close to that of 25 unedited ones in simulation and outperforms them on all four real-robot tasks. %
\end{itemize}

\section{Related Work}
\label{sec:related}

\paragraph{Generating data from demonstrations.}
MimicGen and its successors~\cite{mandlekar2023mimicgen,garrett2025skillmimicgen,jiang2025dexmimicgen} split several annotated demonstrations into object-centric segments, transform each segment to a new object pose, and keep the replays that succeed in a simulator.
Image-based methods instead inpaint objects and backgrounds~\cite{yu2023scaling,chen2023genaug,mandi2023cacti}, synthesize new views~\cite{liu2025dcoda}, or re-render the scene from 3D Gaussians~\cite{yang2025novel,yu2025real2render2real,hu2026manisplat}.
Beyond meta-learning~\cite{duan2017oneshot} and trajectory retrieval~\cite{dipalo2024dinobot}, one-shot editors turn one demonstration into many in 3D.
DemoGen~\cite{xue2025demogen} segments the objects in the point cloud and transforms each object and the matching part of the hand trajectory with a rigid pose.
Its successors~\cite{xu2026r2rgen,zhou2025you,zhou2026oneshot,jiang2026task,ameperosa2024rocoda,xu2026r2rdreamer,liu2026umr,qian2026onedemo} add bimanual coordination, recombined task components, video-completed RGB views, per-object SE(3) trajectories, or hand trajectories edited by reinforcement learning.
These editors work well for new placements, but they move each object as one rigid body.
For deformable objects, DeformGen~\cite{lin2026deformgen} simulates plausible new states and warps the source trajectory, and SoftMimicGen~\cite{moghani2026softmimicgen} executes adapted trajectories in simulation and keeps the successful ones.
In contrast, \method edits the recorded 3D flow point by point, so one kind of operation covers new placements, another drawer, and new rope shapes without simulating or rendering each sample. %

\paragraph{Generating data with language models.}
Instead of starting from demonstrations, GenSim, GenSim2, RoboGen, and Gen2Sim~\cite{wang2024gensim,hua2024gensim2,wang2024robogen,katara2023gen2sim} let language models propose tasks, build simulated scenes, and write rewards or solution code, and then collect many demonstrations in simulation.
Code as Policies~\cite{liang2023code}, VoxPoser, ReKep, and MOKA~\cite{huang2023voxposer,huang2025rekep,fang2024moka} likewise let language models write programs or constraints, from which a planner computes the motion at test time.
In these methods, the motion comes from a script, a planner, or reinforcement learning.
\method instead starts from one demonstration: a VLM names the targets of the edits and writes a program of geometric checks that bounds the pose edit, while the motion comes from the demonstrated flow (\cref{sec:method:vlm}).

\paragraph{3D flow and point tracks for policy learning.}
Point tracks and 3D flow describe how a scene moves without an object model, and robot learning uses them in two roles.
First, they serve as a plan or as the state of a world model.
Policies condition on predicted 2D point tracks~\cite{wen2024anypoint,xu2025flow,bharadhwaj2024track2act}, derive the actions from predicted 3D flow~\cite{eisner2022flowbot3d,yuan2025general,fang2026pointpilot}, or predict 3D flow with world models, from generated videos, or with the actions~\cite{zhi20253dflowaction,guo2025flowdreamer,dharmarajan2025dream2flow,collins2026flomo}.
PEWAM~\cite{hu2026keep} conditions the actions on 3D keypoint trajectories compiled from one demonstration and edits them at test time, not to create training data. %
Second, the flow strengthens the features of a visuomotor policy.
PPI and RoTri-Diff~\cite{yang2025gripper,chen2026rotridiff} predict object point flow as an intermediate condition, other policies generate it jointly with the actions~\cite{noh20253d,xiao2026jamb}, and vision-language-action models use it as an expert or auxiliary signal~\cite{wang2026lamp,go2026flowvla}.
These policies improve the action by using the flow as a plan or as a stronger feature, but their action heads also read the observation features, so the action need not stay consistent with the predicted flow (\cref{sec:exp:follow}).
They also learn the flow only from collected demonstrations and do not edit it to create new ones.
In \method, the action generator reads only the flow and the robot state, so its action follows a changed flow more closely, and an edit of the flow creates a consistent training sample for both generators. %

\section{Method}
\label{sec:method}

\begin{figure*}[t]
\centering
\includegraphics[width=\linewidth]{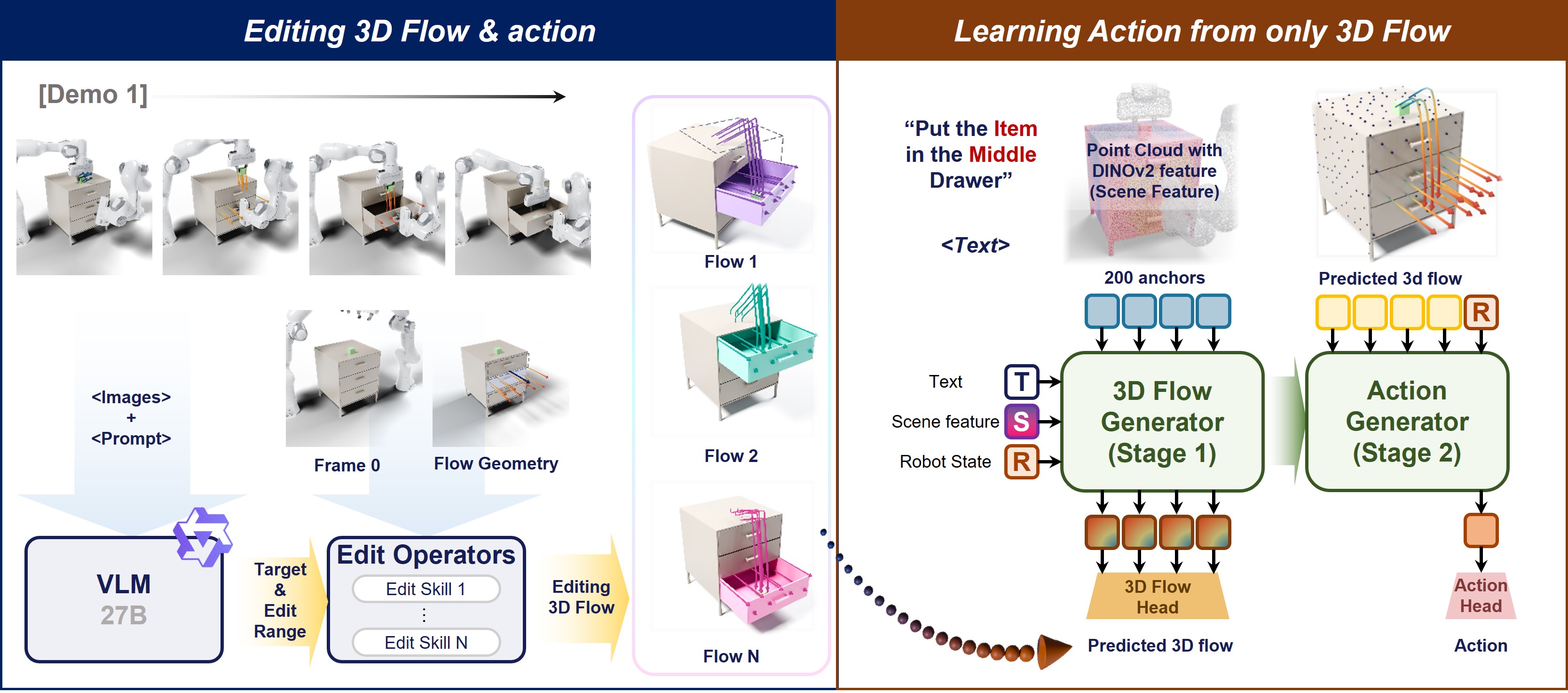}
\caption{\textbf{Edit skills change the 3D flow and the action of one demonstration to train the 3D flow generator and the action generator.}
(Left) Each edit yields a flow label, with edited points, flow, or colors, and, except for the color edit, an action label with the edited action. A VLM names their targets and bounds the pose edit (VLM range). %
(Right) The 3D flow generator predicts the flow from the point cloud, the instruction, and the robot state. The action generator decodes the action from only this flow and the robot state.}
\label{fig:pipeline}
\end{figure*}

\subsection{Problem Statement}
\label{sec:method:problem}

Given one demonstration of a bimanual task with two parallel-jaw grippers, the \textit{source demonstration}, and its language instruction $\ell$, we learn a policy $\pi: (P_t, s_t, \ell) \mapsto A_t$ that also succeeds in unseen scene states, such as new object placements, newly instructed targets, and new start shapes of a deformable object.
Here, $P_t \in \mathbb{R}^{N\times 3}$ is the point cloud fused from calibrated RGB-D cameras at time $t$, $s_t$ is the robot state, and $A_t = (a_t, \ldots, a_{t+K-1})$ is a chunk of $K$ actions, each the target poses of both hands and the target states of both grippers. %
The 3D flow $F_t \in \mathbb{R}^{M\times |\mathcal{H}| \times 3}$ stores the positions of $M$ anchor points $Q_t$, sampled from the non-robot points of $P_t$, at the future offsets $\mathcal{H} = \{0, 25, 50, 75, 100\}$ steps; the anchors need not belong to one rigid body.
The flow thus describes what should happen to the scene, not how the robot should move. %
An edit skill $E_\phi$ with argument $\phi$ maps the demonstration $\mathcal{D} = \{(P_t, s_t, F_t, A_t)\}_{t=1}^{T}$ and $\ell$ to a new demonstration $(\mathcal{D}', \ell')$ whose points, flow, and actions are consistent.
The arguments that need more than geometry, the name of a new target and how far the scene may move, are set by a VLM once per task, offline, and the policy never calls it (\cref{sec:method:vlm}). %

\subsection{Editing 3D Flow and Action}
\label{sec:method:edit}

\begin{figure*}[t]
\centering
\includegraphics[width=0.8\linewidth]{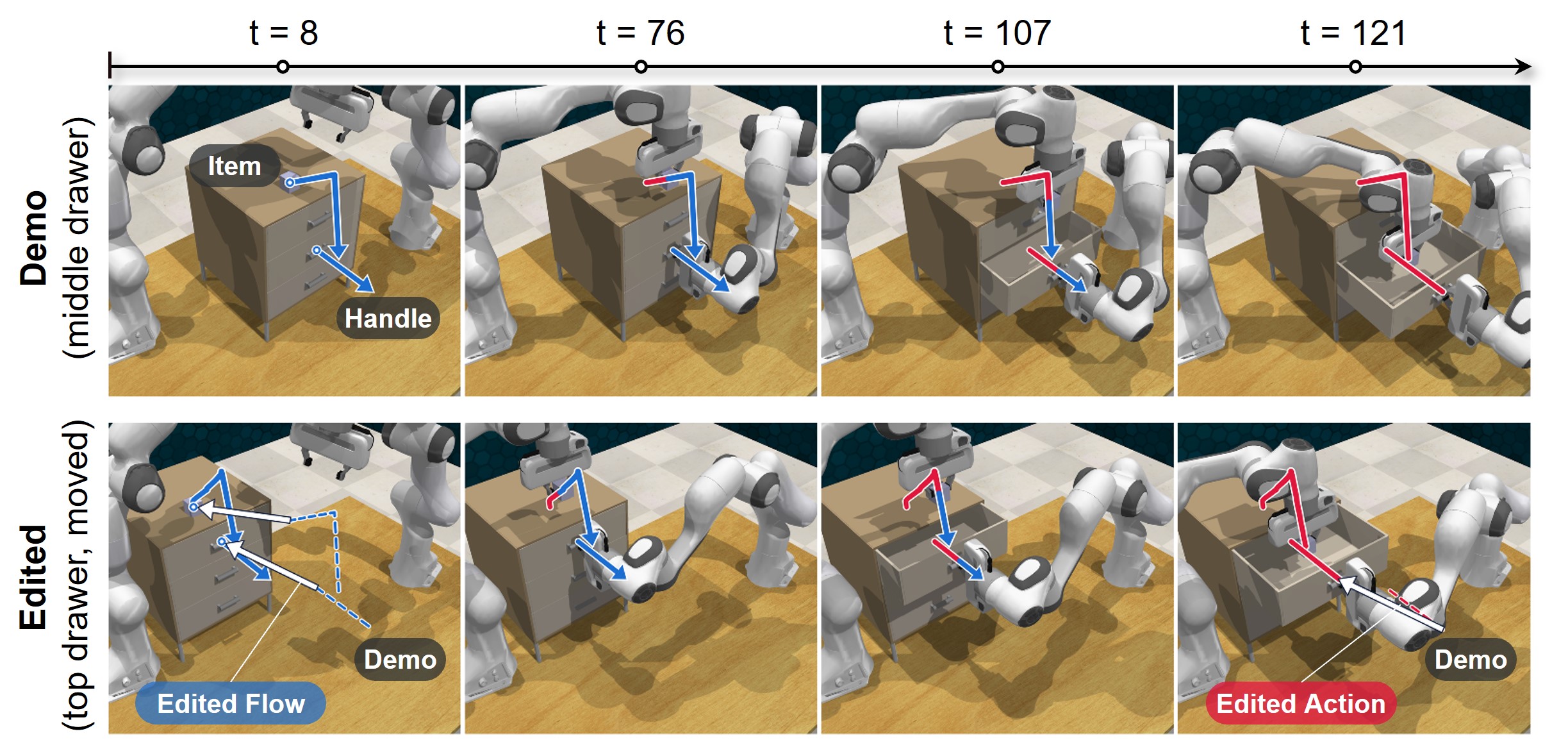}
\caption{\textbf{One edit changes the flow and the action together.}
Top: the demonstration of put item in drawer (middle drawer).
Bottom: an edited training sample, in which the item goes into the top drawer and the whole scene is moved.
Blue: 3D flow of the item and the handle; red: hand actions; dashed lines and white arrows: the demonstration and the displacement from it.
The bottom row is a training label with arms posed by inverse kinematics, not a rollout; the edit is exaggerated for illustration.} %
\label{fig:edit_drawer}
\end{figure*}

\begin{figure*}[t]
\centering
\includegraphics[width=0.75\linewidth]{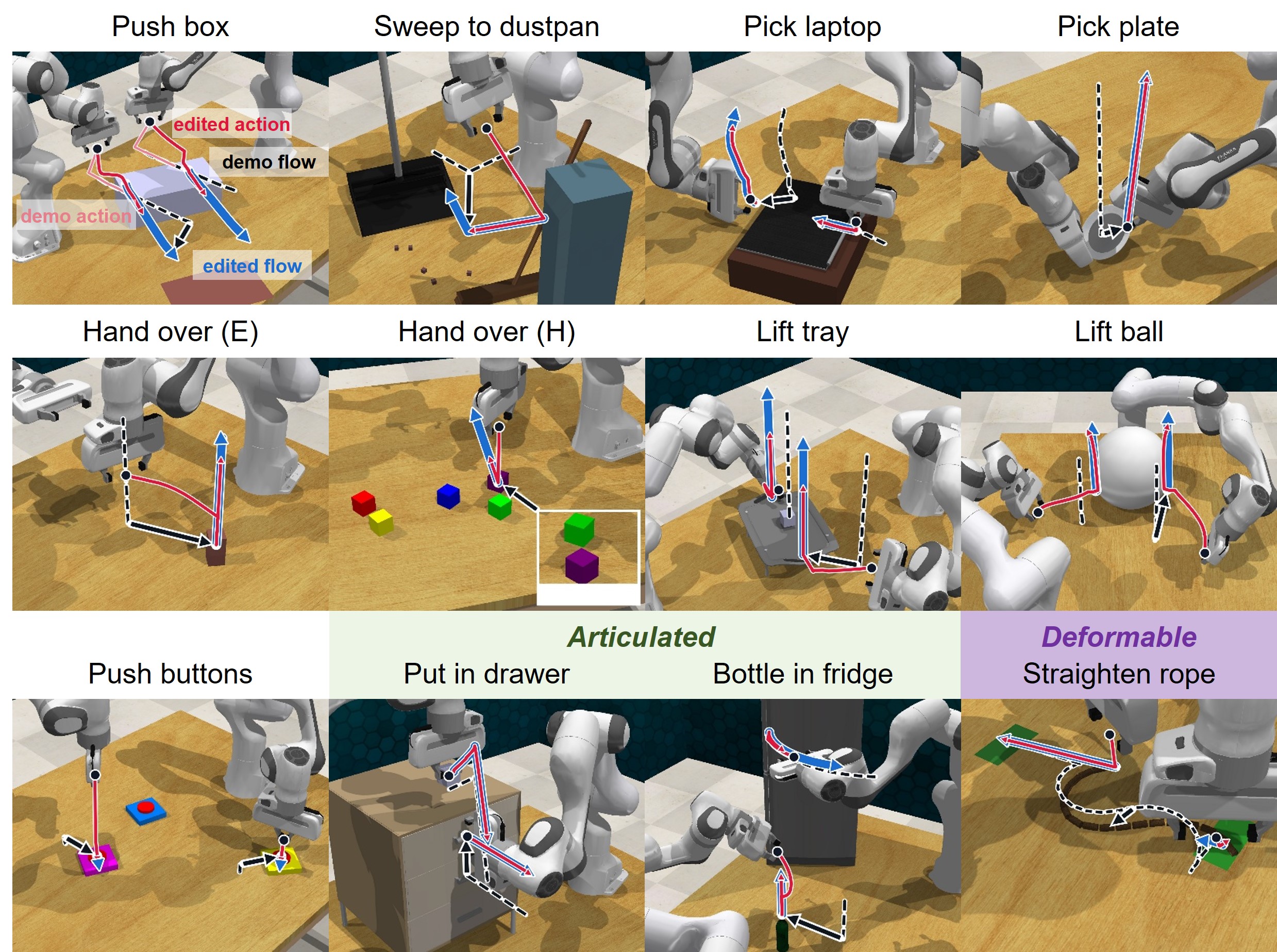} %
\caption{\textbf{Edited labels on all 12 tasks.}
Each panel shows one edit of the source demonstration: demonstrated flow (black, dashed), edited flow (blue), demonstrated action (light red), and edited action (red).
Labels are computed by the edit skills, not policy rollouts; for illustration, the press in push buttons is lengthened and some colors are altered.} %
\label{fig:edit_all}
\end{figure*}

We define four edit skills by the type of object change rather than by the task: a rigid placement (pose), an appearance (color), an articulated part (articulation), and a deformable shape (deformation).
Each task uses the skills that apply to its objects.
Except for the color edit, which rewrites only the point features and the instruction, a skill moves the points, their flow, and the hand targets as a displacement field (\cref{fig:edit_drawer,fig:edit_all}): each anchor trajectory $x_i(t)$ becomes $x'_i(t) = T_{i,t}\big(x_i(t)\big)$, where $T_{i,t}$ is one rigid transform in the pose edit and varies with the anchor and the time in the articulation and deformation edits.
Since the demonstration stores a flow at every time step, an edit of the start state carries through the whole motion consistently. %

Each edit yields two labels for the two networks of the policy (\cref{sec:method:policy}).
The \textit{flow label} $(P', s', F', \ell')$ trains the 3D flow generator: the edited points, their flow, and, if the target changes, the new instruction.
The \textit{action label} $(F', s', A')$ trains the action generator: the edited action on the edited flow.
Each network draws its own applicable edit per training sample.

\paragraph{Pose edit.}
The argument is a rigid scene motion $(\theta, \Delta x, \Delta y)$: the non-robot points rotate by $\theta$ about the vertical axis through their center and translate by $(\Delta x, \Delta y)$, and the flow moves with them.
For the flow label, the motion is drawn from a fixed range that covers the test scenes, because the flow generator only has to recognize a moved scene (\cref{sec:supp:method_details}).
For the action label, the motion is kept only if the arms can still execute it, as decided by the checks that the VLM selects for the task, the \textit{VLM range} (\cref{sec:method:vlm}).
Before contact, the anchors and their flow move while the hands stay, and each hand's target trajectory then blends into the moved scene, at the latest by the first contact.

\paragraph{Color edit.}
The color edit changes which of several items is the target; in handover item (hard), the instruction names one of five item colors.
The skill swaps the RGB values and point features of the demonstrated item and another item.
The instruction then names the new color of the demonstrated item by its VLM label.
Unlike appearance randomization, the edit keeps every color in the scene, and the positions, flow, and actions stay the same.
Since the action generator learns only from the flow and the robot state, the color edit yields no action label and trains only the flow generator. %

\paragraph{Articulation edit.}
The articulation edit transplants the demonstrated opening motion to another drawer $u$ of the same cabinet.
Registering the two drawers in the first frame gives the translation $\Delta_u$ between them, and moving their points and flow by $\Delta_u$ and $-\Delta_u$ exchanges their places.
Drawer $u$ thus opens with the flow of the demonstrated drawer, although its own motion is never observed; the opening direction comes from the flow, so no joint axis or articulation model is estimated. %
The item and the hands follow $\Delta_u$ (\cref{sec:supp:method_details}), and the instruction names drawer $u$ by its VLM label.

\paragraph{Deformation edit.}
The deformation edit changes the start shape of the rope in straighten rope and keeps the demonstrated goal shape.
A displacement $d$ of the rope's nodes fades out as the ends of the rope move,
\begin{equation}
  X'(t) = X(t) + \big(1 - \sigma(t)\big)\, d,
  \label{eq:rope}
\end{equation}
where $X(t)$ denotes the node positions at time $t$ and $\sigma(t) \in [0, 1]$ is the mean motion progress of the two end nodes.
We project the result so that the rope keeps its length with the grasped ends fixed.
The rope points, the nearby robot points, and the hand poses follow the edited rope in both the flow and the action.

\subsection{Grounding Skill Arguments with a VLM}
\label{sec:method:vlm}

The VLM, Qwen3.5-27B~\cite{qwen2026qwen35}, names the edit targets and bounds the pose edit from the first frame of one demonstration of the task (\cref{sec:supp:vlm,sec:supp:vlm_prompts}). %

\paragraph{Target names.}
The color and articulation edits need the name of the new target in the instruction, but the demonstration names only its original one.
The VLM ties each target word to an object found by geometry.
It locates the five color words in handover item (hard) and tells which of the two other drawers is the higher one in put item in drawer (\cref{sec:supp:vlm}).

\paragraph{VLM range.}
How far a scene can be moved depends on the task, because the arms must still reach every grasped or pushed point.
From a top-down rendering of the first frame, marked with the arm bases and the demonstrated hand positions, the VLM writes a function \texttt{allowed}$(\theta, \Delta x, \Delta y)$ as a program, as in Code as Policies~\cite{liang2023code}, over four checks computed with inverse kinematics and scene geometry.
The final functions keep only the reach check on 11 of the 12 tasks and add an on-table check for push box (\cref{sec:supp:vlm}). %

\subsection{Learning Action from Only 3D Flow}
\label{sec:method:policy}

\paragraph{3D flow generator.}
The flow generator $\mathcal{G}$ predicts $\hat F_t = \mathcal{G}(P_t, s_t, \ell)$.
It uses the observation encoder of PPI~\cite{yang2025gripper}, which embeds the $N{=}6144$ points with DINOv2 features~\cite{oquab2024dinov2}, the instruction, and the state.
A Transformer decoder~\cite{vaswani2017attention} with $M{=}200$ anchor queries then predicts each anchor's position at every offset in $\mathcal{H}$.
We train it with an $L_1$ loss against the edited ground-truth flow.
On the tasks whose instruction names the target, a contrastive term reruns the flow generator with another target in the instruction and separates the flows of the two targets (\cref{sec:supp:method_details,sec:supp:real}). %

\paragraph{Action generator.}
The action generator $\mathcal{A}$ predicts $\hat A_t = \mathcal{A}(\hat F_t, s_t)$ and learns only from the flow and the robot state.
Each anchor becomes one token that holds its position relative to the scene center and its displacements at the four future offsets,
\begin{equation}
  z_i = \big[\, q_i - c,\; \{ 10\,\kappa\,(f_{i,h} - f_{i,0}) \}_{h \in \mathcal{H}\setminus\{0\}} \,\big] \in \mathbb{R}^{15},
  \label{eq:anchor_token}
\end{equation}
where $q_i \in Q_t$ is the observed position of anchor $i$, $f_{i,h}$ is its position at offset $h$ in the flow, $c$ is the mean of the non-robot points, and the motion scale $\kappa$ enlarges small motions, such as pressing buttons (\cref{sec:supp:method_details}).
Each anchor token also embeds its offsets to both hands, in the scene frame and in each hand's frame, through a small MLP.
A Transformer encoder reads a state token and the $M$ anchor tokens.
The output at the state token predicts a chunk of $K{=}50$ actions~\cite{zhao2023learning} relative to the current hand poses.
The output is a mixture of three chunks~\cite{bishop1994mixture}; we execute the first ten actions of the mean of the most likely chunk, and then both generators run again.

\paragraph{Training the action generator on ground-truth flow.}
We train the action generator on the edited ground-truth flow, so it learns how a given scene motion is executed, regardless of how the motion was predicted.
We perturb this flow with errors matched to the flow generator on held-out demonstrations and shift it in time (\textit{temporal shift}).
We also keep a random subset of the anchors (\textit{anchor subset}), rotate and translate the scene with the hands (\textit{rigid scene transform}), and perturb the hand poses in the state so that it learns to recover from small deviations~\cite{laskey2017dart}. %

\section{Experiments}
\label{sec:experiments}

\subsection{Setup}
\label{sec:exp:setup}

\paragraph{Datasets.}
We evaluate \method on 12 bimanual tasks of RLBench2~\cite{james2020rlbench,grotz2024peract2} (\cref{tab:oneshot}).
In the one-shot setting, each task is learned from its source demonstration.
We test on 50 test scenes, held-out placements of the source scene with new targets and rope shapes.
They rotate the scene by up to $30^\circ$ and translate it by up to 20\,cm, inside the range of the pose edit and, for nearly all scenes, inside the VLM range (\cref{sec:supp:protocol,sec:supp:vlm}).
Besides the source demonstration, the one-shot setting reads a motion scale and a moving-anchor threshold from the 80 training demonstrations, and two error deviations fitted to each row's own flow generator on the 20 held-out demonstrations; neither generator trains on these demonstrations, and both rows use the same calibration data and procedure (\cref{sec:supp:protocol}). %
In the full-data setting, as in PPI and RoTri-Diff, we train on 80 demonstrations of each task and test on the official split of 100 episodes.
We report the success rate of each task and its mean over the tasks; standard deviations are over three training seeds (details in \cref{sec:supp:protocol,sec:supp:selection}).

\paragraph{Baselines.}
In the one-shot setting, we compare \method with the same policy trained on the source demonstration without edits (1 demo), which also omits the flow augmentations of the action generator (\cref{sec:supp:protocol}). %
We also compare with expert demonstrations of the same task (\cref{sec:exp:demos}) and with a single-stage model trained on the one-shot edits that apply to it (\cref{sec:exp:follow}).
In the full-data setting, we compare with \method without edits and with the values that Table~II of RoTri-Diff~\cite{chen2026rotridiff} reports for RoTri-Diff, PPI~\cite{yang2025gripper}, and three other policies, whose protocols differ from ours (\cref{sec:supp:protocol}). %
In full-data simulation, \method uses only the flow label of the pose edit; the other edits are used in one-shot simulation, and the real robot uses the edits that apply to its tasks (\cref{sec:supp:real}).

\begin{table*}[t]
\centering
\scriptsize
\setlength{\tabcolsep}{3pt}
\def\pmsd#1#2{#1{\tiny$\pm$#2}}
\caption{\textbf{Flow edits improve one-shot success across 12 RLBench2 tasks.} Success rate (\%) on the 50 test scenes per task, mean $\pm$ standard deviation over three seeds. Both rows train the same policy from the source demonstration.} %
\label{tab:oneshot}
\begin{tabular}{l*{13}{c}}
\hline
 & \shortstack{push\\box} & \shortstack{push\\buttons} & \shortstack{lift\\tray} & \shortstack{pick\\laptop} & \shortstack{lift\\ball} & \shortstack{handover\\easy} & \shortstack{handover\\hard} & \shortstack{put in\\drawer} & \shortstack{sweep to\\dustpan} & \shortstack{bottle in\\fridge} & \shortstack{pick\\plate} & \shortstack{straighten\\rope} & Mean \\
\hline
1 demo & \pmsd{\OneShotBaseBox}{\OneShotBaseBoxSd} & \pmsd{\OneShotBaseButtons}{\OneShotBaseButtonsSd} & \pmsd{\OneShotBaseTray}{\OneShotBaseTraySd} & \pmsd{\OneShotBaseLaptop}{\OneShotBaseLaptopSd} & \pmsd{\OneShotBaseBall}{\OneShotBaseBallSd} & \pmsd{\OneShotBaseHandEasy}{\OneShotBaseHandEasySd} & \pmsd{\OneShotBaseHandHard}{\OneShotBaseHandHardSd} & \pmsd{\OneShotBaseDrawer}{\OneShotBaseDrawerSd} & \pmsd{\OneShotBaseDustpan}{\OneShotBaseDustpanSd} & \pmsd{\OneShotBaseFridge}{\OneShotBaseFridgeSd} & \pmsd{\OneShotBasePlate}{\OneShotBasePlateSd} & \pmsd{\OneShotBaseRope}{\OneShotBaseRopeSd} & \pmsd{\OneShotBaseMean}{\OneShotBaseMeanSd} \\
1 demo + edits (\method) & \pmsd{\textbf{\OneShotEditBox}}{\OneShotEditBoxSd} & \pmsd{\textbf{\OneShotEditButtons}}{\OneShotEditButtonsSd} & \pmsd{\textbf{\OneShotEditTray}}{\OneShotEditTraySd} & \pmsd{\textbf{\OneShotEditLaptop}}{\OneShotEditLaptopSd} & \pmsd{\textbf{\OneShotEditBall}}{\OneShotEditBallSd} & \pmsd{\textbf{\OneShotEditHandEasy}}{\OneShotEditHandEasySd} & \pmsd{\textbf{\OneShotEditHandHard}}{\OneShotEditHandHardSd} & \pmsd{\textbf{\OneShotEditDrawer}}{\OneShotEditDrawerSd} & \pmsd{\textbf{\OneShotEditDustpan}}{\OneShotEditDustpanSd} & \pmsd{\textbf{\OneShotEditFridge}}{\OneShotEditFridgeSd} & \pmsd{\textbf{\OneShotEditPlate}}{\OneShotEditPlateSd} & \pmsd{\textbf{\OneShotEditRope}}{\OneShotEditRopeSd} & \pmsd{\textbf{\OneShotEditMean}}{\OneShotEditMeanSd} \\
\hline
\end{tabular}
\end{table*}

\subsection{One-Shot Generalization}
\label{sec:exp:main}

A single edited demonstration raises the average success rate from \OneShotBaseMean\% to \OneShotEditMean\% and, on average over three seeds, improves all 12 tasks (\cref{tab:oneshot}). %
On nine tasks, the edits are only the pose edit; the other three also use the edits of \cref{sec:exp:beyond}.
The pose edit moves the scene as one rigid body, a change that rigid editors such as DemoGen also cover, so \cref{tab:loo} measures what the other point-wise edits add on top of it. %

\begin{figure}[t]
\centering
\includegraphics[width=\columnwidth]{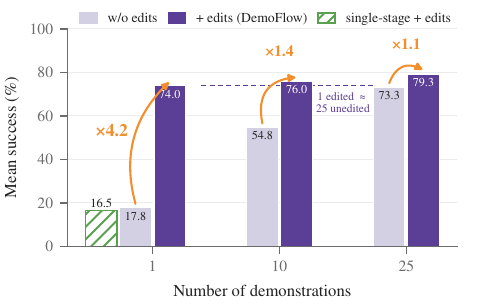}
\caption{\textbf{One edited demonstration is close to 25 unedited demonstrations.} Mean success rate over the 12 tasks on the 50 test scenes of \cref{tab:oneshot}, for one training run. Arrows give the ratio of the edited to the unedited rate. The 10 and 25 expert demonstrations are nested, drawn from the range of the test scenes, and disjoint from them; per-task results are in \cref{tab:supp:demos}.}
\label{fig:demos}
\end{figure}

\subsection{Edits versus More Demonstrations}
\label{sec:exp:demos}

On the 50 test scenes of \cref{tab:oneshot}, averaged over the 12 tasks, one edited demonstration reaches \DemoOneEditMean\%, close to the \DemoTwentyFiveNoEditMean\% of 25 unedited expert demonstrations and above the \DemoTenNoEditMean\% of 10 (\cref{fig:demos}). %
Since the expert demonstrations come from the same range as the test scenes, this comparison asks whether edited labels are as useful as collected ones within that range. %
On average, the edits also improve the rows with 10 and 25 demonstrations, with task-level exceptions. %
With 25 demonstrations, they raise put item in drawer from \DemoTwentyFiveNoEditDrawer\% to \DemoTwentyFiveEditDrawer\% and handover item (hard) from \DemoTwentyFiveNoEditHandHard\% to \DemoTwentyFiveEditHandHard\% but lower lift tray, pick laptop, and lift ball (\cref{tab:supp:demos}). %

\subsection{Edits Beyond Rigid Placement}
\label{sec:exp:beyond}

\begin{table}[t]
\centering
\scriptsize
\setlength{\tabcolsep}{4pt}
\caption{\textbf{Edits beyond the rigid pose edit raise success on their tasks.} Gain in success rate (percentage points) over the pose edit only, on the scenes of \cref{tab:oneshot}; mean $\pm$ standard deviation over three seeds.}
\label{tab:loo}
\begin{tabular}{llc}
\hline
Task & Added edit & Gain \\
\hline
handover item (hard) & color & +\LooDropHandHardAvg{\tiny$\pm$\LooDropHandHardSd} \\
put item in drawer & articulation & +\LooDropDrawerAvg{\tiny$\pm$\LooDropDrawerSd} \\
straighten rope & deformation & +\LooDropRopeAvg{\tiny$\pm$\LooDropRopeSd} \\
\hline
\end{tabular}
\end{table}

Averaged over three seeds, the color edit raises handover item (hard) from \LooWithoutHandHardAvg\% with the pose edit alone to \OneShotEditHandHard\%, and the articulation edit raises put item in drawer from \LooWithoutDrawerAvg\% to \OneShotEditDrawer\%, with a gain in every seed.
The deformation edit raises straighten rope from \LooWithoutRopeAvg\% to \OneShotEditRope\%, with no gain in one of the three seeds. %

\subsection{Action from Only 3D Flow}
\label{sec:exp:follow}

\begin{table}[t]
\centering
\scriptsize
\setlength{\tabcolsep}{4pt}
\caption{\textbf{Changing the flow changes the action of \method more than that of PPI.} At the first query of an episode, we rotate by $15^\circ$ the flow of \method or the object points from which PPI predicts its flow, or switch the drawer instruction for both. Follow ratio $r$: the shift of the hand target toward the target implied by the change, as a fraction of the full shift (1: the full shift, 0: no shift, below 0: a shift the other way). Mean over 20 test episodes; all interventions are in \cref{tab:follow_full}.} %
\label{tab:follow}
\begin{tabular}{llcc}
\hline
Task & Intervention & PPI & \method \\
\hline
lift tray & rotation & \FollowPpiTrayRot & \textbf{\FollowOursTrayRot} \\
push box & rotation & \FollowPpiBoxRot & \textbf{\FollowOursBoxRot} \\
sweep to dustpan & rotation & $\FollowPpiDustpanRot$ & \textbf{\FollowOursDustpanRot} \\
pick laptop & rotation & \FollowPpiLaptopRot & \textbf{\FollowOursLaptopRot} \\
put in drawer & instruction & \FollowPpiDrawerInstr & \textbf{\FollowOursDrawerInstr} \\
\hline
\end{tabular}
\end{table}

\paragraph{Does the action follow the flow?}
In all five interventions, the action of \method follows a changed flow more than that of PPI, most clearly under another instruction and on sweep to dustpan (\cref{tab:follow}). %
We intervene on the full-data models; PPI predicts its flow inside the network, and its action head has no separate flow input (\cref{sec:supp:interventions}). %
Under rotation, the predicted flow of PPI rotates on four tasks, but its action follows the rotation less, consistent with its flow strengthening the observation features rather than deciding the action.
Under another instruction, its predicted flow stays the same, and its hand does not move toward the other drawer. %
In closed loop, the action generator follows even a rotated flow that does not fit the scene (\cref{sec:supp:interventions}), so its success falls from \FollowOursSuccBaseLaptop\% to \FollowOursSuccRotLaptop\% on pick laptop and from 100\% to \FollowOursSuccRotBox\% on push box, %
against \FollowPpiSuccRotLaptop\% and \FollowPpiSuccRotBox\% for PPI.

\paragraph{Can a single network replace the two stages?}
We test a single-stage model that attaches the decoder of the action generator to the flow generator and keeps the flow only as an auxiliary output.
With a lighter trunk and only the edits that apply to a single network, it reaches \OneStageMean\% (\cref{fig:demos}).
It is below \method on every one of the 12 tasks and near the \DemoOneNoEditMean\% of \method without edits (\cref{sec:supp:single}). %
\subsection{All Training Demonstrations}
\label{sec:exp:full}

\begin{table*}[t]
\centering
\scriptsize
\setlength{\tabcolsep}{3pt}
\def\pmsd#1#2{#1{\tiny$\pm$#2}}
\caption{\textbf{With all training demonstrations, \method is competitive with reported flow-conditioned policies on RLBench2.} Success rate (\%) on 100 test episodes per task; Avg.\ success is over the 11 tasks in Table~II of RoTri-Diff~\cite{chen2026rotridiff}, which omits straighten rope. Rows above the line are reported values under different protocols; the PPI result on pick plate is obtained without re-seating the arms after reset (\cref{sec:supp:protocol}). RoTri-Diff and \method: mean $\pm$ standard deviation over three seeds; w/o edits: no edit-related augmentation.} %
\label{tab:full}
\begin{tabular*}{\textwidth}{@{\extracolsep{\fill}}l*{13}{c}}
\hline
 & \shortstack{Avg.\\success} & \shortstack{push\\buttons} & \shortstack{lift\\ball} & \shortstack{lift\\tray} & \shortstack{push\\box} & \shortstack{handover\\easy} & \shortstack{handover\\hard} & \shortstack{pick\\laptop} & \shortstack{put in\\drawer} & \shortstack{sweep to\\dustpan} & \shortstack{pick\\plate} & \shortstack{bottle in\\fridge} & \shortstack{straighten\\rope} \\
\hline
ACT~\cite{zhao2023learning} & \FullActMean & \FullActButtons & \FullActBall & \FullActTray & \FullActBox & \FullActHandEasy & \FullActHandHard & \FullActLaptop & \FullActDrawer & \FullActDustpan & \FullActPlate & \FullActFridge & -- \\
PerAct$^2$~\cite{grotz2024peract2} & \FullPeractMean & \FullPeractButtons & \FullPeractBall & \FullPeractTray & \FullPeractBox & \FullPeractHandEasy & \FullPeractHandHard & \FullPeractLaptop & \FullPeractDrawer & \FullPeractDustpan & \FullPeractPlate & \FullPeractFridge & -- \\
AnyBimanual~\cite{lu2025anybimanual} & \FullAnyBiMean & \FullAnyBiButtons & \FullAnyBiBall & \FullAnyBiTray & \FullAnyBiBox & \FullAnyBiHandEasy & \FullAnyBiHandHard & \FullAnyBiLaptop & \FullAnyBiDrawer & \FullAnyBiDustpan & \FullAnyBiPlate & \FullAnyBiFridge & -- \\
PPI~\cite{yang2025gripper} & \FullPpiMean & \FullPpiButtons & \FullPpiBall & \FullPpiTray & \FullPpiBox & \FullPpiHandEasy & \FullPpiHandHard & \FullPpiLaptop & \FullPpiDrawer & \FullPpiDustpan & \FullPpiPlate & \FullPpiFridge & -- \\
RoTri-Diff~\cite{chen2026rotridiff} & \FullRotriMean & \pmsd{\textbf{\FullRotriButtons}}{\FullRotriButtonsSd} & \pmsd{\FullRotriBall}{\FullRotriBallSd} & \pmsd{\FullRotriTray}{\FullRotriTraySd} & \pmsd{\FullRotriBox}{\FullRotriBoxSd} & \pmsd{\textbf{\FullRotriHandEasy}}{\FullRotriHandEasySd} & \pmsd{\FullRotriHandHard}{\FullRotriHandHardSd} & \pmsd{\FullRotriLaptop}{\FullRotriLaptopSd} & \pmsd{\FullRotriDrawer}{\FullRotriDrawerSd} & \pmsd{\FullRotriDustpan}{\FullRotriDustpanSd} & \pmsd{\FullRotriPlate}{\FullRotriPlateSd} & \pmsd{\FullRotriFridge}{\FullRotriFridgeSd} & -- \\
\hline
\method w/o edits & \FullNoEditMean & \FullNoEditButtons & \FullNoEditBall & \textbf{\FullNoEditTray} & \textbf{\FullNoEditBox} & \FullNoEditHandEasy & \FullNoEditHandHard & \FullNoEditLaptop & \textbf{\FullNoEditDrawer} & \FullNoEditDustpan & \FullNoEditPlate & \textbf{\FullNoEditFridge} & \FullNoEditRope \\
\method & \pmsd{\FullOursAvgMean}{\FullOursAvgMeanSd} & \pmsd{\FullOursAvgButtons}{\FullOursAvgButtonsSd} & \pmsd{\textbf{\FullOursAvgBall}}{\FullOursAvgBallSd} & \pmsd{\FullOursAvgTray}{\FullOursAvgTraySd} & \pmsd{\textbf{\FullOursAvgBox}}{\FullOursAvgBoxSd} & \pmsd{\FullOursAvgHandEasy}{\FullOursAvgHandEasySd} & \pmsd{\textbf{\FullOursAvgHandHard}}{\FullOursAvgHandHardSd} & \pmsd{\textbf{\FullOursAvgLaptop}}{\FullOursAvgLaptopSd} & \pmsd{\FullOursAvgDrawer}{\FullOursAvgDrawerSd} & \pmsd{\textbf{\FullOursAvgDustpan}}{\FullOursAvgDustpanSd} & \pmsd{\textbf{\FullOursAvgPlate}}{\FullOursAvgPlateSd} & \pmsd{\FullOursAvgFridge}{\FullOursAvgFridgeSd} & \pmsd{\textbf{\FullOursAvgRope}}{\FullOursAvgRopeSd} \\
\hline
\end{tabular*}
\end{table*}

With all training demonstrations, \method reaches \FullOursAvgMean\% over three seeds on 11 tasks, against the \FullRotriMean\% and \FullPpiMean\% reported for RoTri-Diff and PPI under different protocols (\cref{tab:full}).
Without edits, the same policy reaches \FullNoEditMean\%.
On straighten rope, the edits raise \method from \FullNoEditRope\% to \FullOursAvgRope\%. %

\subsection{Ablation Studies}
\label{sec:exp:ablation}
\label{sec:exp:roles}
The VLM range performs on par with a range computed with inverse kinematics on the same test scenes, \OneShotEditMean\% against \RangeIkAvgMean\% (\cref{tab:range}).
The benefit of the VLM range is generality rather than success: an IK range checks only reachability, whereas the VLM decides which checks the task needs (\cref{sec:method:vlm}). %
As the test scene moves farther from the source scene, both rows of \cref{tab:oneshot} lose success, but the gain of the edits stays between \MagShiftCGain\ and \MagThetaCGain\ points (\cref{tab:magnitude}).

\subsection{Real-World Experiments}
\label{sec:exp:real}

\begin{table}[t]
\centering
\scriptsize
\setlength{\tabcolsep}{3.5pt}
\caption{\textbf{One edited real demonstration outperforms 25 unedited ones on all four tasks.} Success rate (\%) over 25 trials of one trained model per cell. PPI rows train on the same 50 demonstrations as \method, and PPI + edits trains for as many epochs as PPI.}
\label{tab:real}
\begin{tabular}{l*{5}{c}}
\hline
Training data & \shortstack{hang\\cable} & \shortstack{put item\\in drawer} & \shortstack{disassemble\\Lego} & \shortstack{insert\\plate} & Mean \\
\hline
1 demo & \RealNoEditOneCable & \RealNoEditOneDrawer & \RealNoEditOneLego & \RealNoEditOnePlate & \RealNoEditOneMean \\
10 demos & \RealNoEditTenCable & \RealNoEditTenDrawer & \RealNoEditTenLego & \RealNoEditTenPlate & \RealNoEditTenMean \\
25 demos & \RealNoEditTwentyFiveCable & \RealNoEditTwentyFiveDrawer & \RealNoEditTwentyFiveLego & \RealNoEditTwentyFivePlate & \RealNoEditTwentyFiveMean \\
50 demos & \RealNoEditFiftyCable & \RealNoEditFiftyDrawer & \RealNoEditFiftyLego & \RealNoEditFiftyPlate & \RealNoEditFiftyMean \\
\hline
1 demo + edits (\method) & \RealEditOneCable & \RealEditOneDrawer & \RealEditOneLego & \RealEditOnePlate & \RealEditOneMean \\
10 demos + edits & \RealEditTenCable & \RealEditTenDrawer & \RealEditTenLego & \RealEditTenPlate & \RealEditTenMean \\
25 demos + edits & \textbf{\RealEditTwentyFiveCable} & \textbf{\RealEditTwentyFiveDrawer} & \textbf{\RealEditTwentyFiveLego} & \textbf{\RealEditTwentyFivePlate} & \textbf{\RealEditTwentyFiveMean} \\
50 demos + edits & \RealEditFiftyCable & \textbf{\RealEditFiftyDrawer} & \RealEditFiftyLego & \textbf{\RealEditFiftyPlate} & \RealEditFiftyMean \\
\hline
PPI~\cite{yang2025gripper}, 50 demos & \RealPpiFiftyCable & \RealPpiFiftyDrawer & \RealPpiFiftyLego & \RealPpiFiftyPlate & \RealPpiFiftyMean \\
PPI + edits, 50 demos & \RealPpiEditFiftyCable & \RealPpiEditFiftyDrawer & \RealPpiEditFiftyLego & \RealPpiEditFiftyPlate & \RealPpiEditFiftyMean \\
\hline
\end{tabular}
\end{table}

\paragraph{Setup.}
Two Franka Panda arms are teleoperated with the force-feedback leader--follower system of FACTR~\cite{liu2025factr}, and four calibrated RGB-D cameras observe the scene (\cref{fig:real} in the appendix).
The flow labels of a real demonstration come from CoTracker3~\cite{karaev2024cotracker3} tracks lifted to 3D with the aligned depth (\cref{sec:supp:real}).
We design four tasks (\cref{fig:teaser}, right; a demonstration of each and its edits are in \cref{sec:supp:real}).
In hang cable, one hand pulls a deformable cable and the other hangs the ring at its end; in put item in drawer, the robot opens a drawer and puts an item in it; in disassemble Lego, it pulls apart the instructed one of four colored brick pairs; in insert plate, it picks up a plate with a non-prehensile motion and inserts it into a rack.
We train with 1, 10, 25, or 50 of 50 collected demonstrations.
Test scenes are laid out separately from the demonstrations: the drawer test scenes instruct two drawers that the source demonstration never opens, and in \RealYawOutside{} of the \RealYawTotal{} scenes the planned cabinet orientation lies outside those of all 50 collected demonstrations (\cref{sec:supp:real}).

\paragraph{Results.}
With the edited labels, and for disassemble Lego also the contrastive term, one edited demonstration outperforms 25 unedited demonstrations on all four tasks, with an average success rate of \RealEditOneMean\% against \RealNoEditTwentyFiveMean\% (\cref{tab:real}).
On disassemble Lego and insert plate, it also exceeds 50 unedited demonstrations, \RealEditOneLego\% against \RealNoEditFiftyLego\% and \RealEditOnePlate\% against \RealNoEditFiftyPlate\%.
Without edits, the Lego flow generator learns to move the pair nearest the right hand rather than the instructed one, and more demonstrations strengthen this shortcut; with edits and the contrastive term, the predicted flow moves the instructed pair in nearly every trial (\cref{sec:supp:real}). %
The two PPI rows train for the same \PpiRealEpochs{} epochs; with 50 demonstrations, the edits lower PPI and raise \method on every task, consistent with the flow-following gap in \cref{sec:exp:follow}. %

\section{Conclusion}
\label{sec:conclusion}

We presented \method, which creates demonstrations by point-wise editing of the 3D flow of one demonstration and executes them with an action generator that learns only from the flow and the robot state.
One edited demonstration nears 25 unedited ones on average in simulation and outperforms them on all four real-robot tasks.
On top of the rigid pose edit, the edits that a rigid editor cannot express still raise success substantially (\cref{tab:loo}), and on the real robot, the edits that raise \method lower PPI, consistent with how closely each action follows its flow (\cref{sec:exp:follow}). %
These results suggest that 3D flow can serve not only as a representation for a policy but also as a space in which to create its training data.

\paragraph{Limitations and future work.}
Our simulated test scenes are new placements of the source scene with new targets and rope shapes, nearly all within the VLM range, and the one-shot calibration reads a few scalars from other demonstrations of the task.
The edits keep the objects and grasps of the source demonstration, check physical validity only by keeping the rope length, and assume that parts of the same object share one motion type.
Success stays below 50\% on bottle in fridge and sweep to dustpan (\cref{tab:oneshot}). %
On the real robot, most failures occur at contact-rich grasps, presses, and insertions, and the policy keeps the arm roles of the source demonstration (\cref{sec:supp:roles}). %
Because the edits act on individual points, they could extend to the shape of the object and the grasp on it, which would also open the evaluation to new object instances. %

\section*{Acknowledgments}
This work was supported in part by the Industrial Strategic Technology Development Program (RS-2024-00442029) funded by the Ministry of Trade, Industry and Energy (MOTIE, Korea), in part by the National Research Foundation of Korea (NRF) funded by the Korea government (MSIT) (RS-2024-00464386), and in part by the KIST Institutional Program (26E0051 and 26Z9020).

{
    \small
    \bibliographystyle{ieeenat_fullname}
    \bibliography{ref}
}

\input{sec/appendix_arxiv}

\end{document}

%% file: numbers.tex
\newcommand{\OneShotBaseBox}{63.3}  %
\newcommand{\OneShotBaseBoxSd}{8.1}  %
\newcommand{\OneShotBaseButtons}{28.7}  %
\newcommand{\OneShotBaseButtonsSd}{2.3}  %
\newcommand{\OneShotBaseTray}{35.3}  %
\newcommand{\OneShotBaseTraySd}{13.3}  %
\newcommand{\OneShotBaseLaptop}{10.0}  %
\newcommand{\OneShotBaseLaptopSd}{4.0}  %
\newcommand{\OneShotBaseBall}{48.7}  %
\newcommand{\OneShotBaseBallSd}{6.1}  %
\newcommand{\OneShotBaseHandEasy}{2.7}  %
\newcommand{\OneShotBaseHandEasySd}{3.1}  %
\newcommand{\OneShotBaseHandHard}{0.0}  %
\newcommand{\OneShotBaseHandHardSd}{0.0}  %
\newcommand{\OneShotBaseDrawer}{7.3}  %
\newcommand{\OneShotBaseDrawerSd}{1.2}  %
\newcommand{\OneShotBaseDustpan}{4.7}  %
\newcommand{\OneShotBaseDustpanSd}{4.2}  %
\newcommand{\OneShotBaseFridge}{2.0}  %
\newcommand{\OneShotBaseFridgeSd}{2.0}  %
\newcommand{\OneShotBasePlate}{37.3}  %
\newcommand{\OneShotBasePlateSd}{4.6}  %
\newcommand{\OneShotBaseRope}{13.3}  %
\newcommand{\OneShotBaseRopeSd}{6.4}  %
\newcommand{\OneShotBaseMean}{21.1}  %
\newcommand{\OneShotBaseMeanSd}{3.0}  %
\newcommand{\OneShotEditBox}{96.7}  %
\newcommand{\OneShotEditBoxSd}{1.2}  %
\newcommand{\OneShotEditButtons}{98.0}  %
\newcommand{\OneShotEditButtonsSd}{0.0}  %
\newcommand{\OneShotEditTray}{98.7}  %
\newcommand{\OneShotEditTraySd}{2.3}  %
\newcommand{\OneShotEditLaptop}{83.3}  %
\newcommand{\OneShotEditLaptopSd}{8.1}  %
\newcommand{\OneShotEditBall}{77.3}  %
\newcommand{\OneShotEditBallSd}{1.2}  %
\newcommand{\OneShotEditHandEasy}{70.7}  %
\newcommand{\OneShotEditHandEasySd}{9.0}  %
\newcommand{\OneShotEditHandHard}{62.0}  %
\newcommand{\OneShotEditHandHardSd}{21.6}  %
\newcommand{\OneShotEditDrawer}{67.3}  %
\newcommand{\OneShotEditDrawerSd}{12.2}  %
\newcommand{\OneShotEditDustpan}{46.0}  %
\newcommand{\OneShotEditDustpanSd}{8.7}  %
\newcommand{\OneShotEditFridge}{38.7}  %
\newcommand{\OneShotEditFridgeSd}{11.0}  %
\newcommand{\OneShotEditPlate}{83.3}  %
\newcommand{\OneShotEditPlateSd}{6.1}  %
\newcommand{\OneShotEditRope}{55.3}  %
\newcommand{\OneShotEditRopeSd}{9.5}  %
\newcommand{\OneShotEditMean}{73.1}  %
\newcommand{\OneShotEditMeanSd}{0.9}  %
\newcommand{\FullRotriBox}{95.0}  %
\newcommand{\FullRotriButtons}{97.0}  %
\newcommand{\FullRotriTray}{94.3}  %
\newcommand{\FullRotriLaptop}{66.0}  %
\newcommand{\FullRotriBall}{95.7}  %
\newcommand{\FullRotriHandEasy}{73.3}  %
\newcommand{\FullRotriHandHard}{52.3}  %
\newcommand{\FullRotriDrawer}{87.0}  %
\newcommand{\FullRotriDustpan}{96.7}  %
\newcommand{\FullRotriFridge}{92.0}  %
\newcommand{\FullRotriPlate}{40.7}  %
\newcommand{\FullRotriMean}{80.9}  %
\newcommand{\FullNoEditBox}{99.0}  %
\newcommand{\FullNoEditButtons}{92.0}  %
\newcommand{\FullNoEditTray}{99.0}  %
\newcommand{\FullNoEditLaptop}{67.0}  %
\newcommand{\FullNoEditBall}{96.0}  %
\newcommand{\FullNoEditHandEasy}{62.0}  %
\newcommand{\FullNoEditHandHard}{26.0}  %
\newcommand{\FullNoEditDrawer}{97.0}  %
\newcommand{\FullNoEditDustpan}{79.0}  %
\newcommand{\FullNoEditFridge}{96.0}  %
\newcommand{\FullNoEditPlate}{75.0}  %
\newcommand{\FullNoEditMean}{80.7}  %
\newcommand{\FollowOursDrawerShift}{0.85}  %
\newcommand{\FollowOursDrawerRot}{0.54}  %
\newcommand{\FollowOursDrawerInstr}{1.00}  %
\newcommand{\FollowOursDrawerOther}{1.00}  %
\newcommand{\FollowOursTrayShift}{0.91}  %
\newcommand{\FollowOursTrayRot}{0.55}  %
\newcommand{\FollowOursTrayOther}{1.00}  %
\newcommand{\FollowOursBoxShift}{0.50}  %
\newcommand{\FollowOursBoxRot}{0.59}  %
\newcommand{\FollowOursBoxOther}{1.00}  %
\newcommand{\FollowPpiDrawerShift}{0.77}  %
\newcommand{\FollowPpiDrawerRot}{0.01}  %
\newcommand{\FollowPpiDrawerInstr}{0.00}  %
\newcommand{\FollowPpiDrawerOther}{0.49}  %
\newcommand{\FollowPpiTrayShift}{0.71}  %
\newcommand{\FollowPpiTrayRot}{0.37}  %
\newcommand{\FollowPpiTrayOther}{0.98}  %
\newcommand{\FollowPpiBoxShift}{0.47}  %
\newcommand{\FollowPpiBoxRot}{0.26}  %
\newcommand{\FollowPpiBoxOther}{0.95}  %
\newcommand{\FollowPpiBaseDrawer}{19 / 20}  %
\newcommand{\FollowPpiOwnTrayRot}{1.00}  %
\newcommand{\FollowPpiBaseTray}{18 / 20}  %
\newcommand{\FollowPpiOwnBoxRot}{1.00}  %
\newcommand{\FollowPpiBaseBox}{19 / 20}  %
\newcommand{\MagThetaAN}{218}  %
\newcommand{\MagThetaBN}{211}  %
\newcommand{\MagThetaCN}{171}  %
\newcommand{\MagShiftAN}{169}  %
\newcommand{\MagShiftBN}{338}  %
\newcommand{\MagShiftCN}{93}  %
\newcommand{\MagThetaABase}{29.1}  %
\newcommand{\MagThetaABaseSd}{3.4}  %
\newcommand{\MagThetaAEdit}{77.1}  %
\newcommand{\MagThetaAEditSd}{2.0}  %
\newcommand{\MagThetaAGain}{48.1}  %
\newcommand{\MagThetaBBase}{18.6}  %
\newcommand{\MagThetaBBaseSd}{3.4}  %
\newcommand{\MagThetaBEdit}{71.8}  %
\newcommand{\MagThetaBEditSd}{0.6}  %
\newcommand{\MagThetaBGain}{53.1}  %
\newcommand{\MagThetaCBase}{14.0}  %
\newcommand{\MagThetaCBaseSd}{2.3}  %
\newcommand{\MagThetaCEdit}{69.7}  %
\newcommand{\MagThetaCEditSd}{1.2}  %
\newcommand{\MagThetaCGain}{55.7}  %
\newcommand{\MagShiftABase}{33.7}  %
\newcommand{\MagShiftABaseSd}{2.6}  %
\newcommand{\MagShiftAEdit}{84.4}  %
\newcommand{\MagShiftAEditSd}{3.4}  %
\newcommand{\MagShiftAGain}{50.6}  %
\newcommand{\MagShiftBBase}{17.5}  %
\newcommand{\MagShiftBBaseSd}{4.1}  %
\newcommand{\MagShiftBEdit}{71.5}  %
\newcommand{\MagShiftBEditSd}{0.7}  %
\newcommand{\MagShiftBGain}{54.0}  %
\newcommand{\MagShiftCBase}{11.5}  %
\newcommand{\MagShiftCBaseSd}{2.2}  %
\newcommand{\MagShiftCEdit}{58.6}  %
\newcommand{\MagShiftCEditSd}{4.9}  %
\newcommand{\MagShiftCGain}{47.2}  %
\newcommand{\TimeScene}{3.2}  %
\newcommand{\TimeRope}{2.6}  %
\newcommand{\TimeSim}{350}  %
\newcommand{\SizeOursTotal}{52.7}  %
\newcommand{\SizeOursTrain}{11.1}  %
\newcommand{\SizeOursCall}{105 ms}  %
\newcommand{\SizeGenTotal}{47.4}  %
\newcommand{\SizeGenTrain}{5.8}  %
\newcommand{\SizeGenCall}{76 ms}  %
\newcommand{\SizeReaderTotal}{5.3}  %
\newcommand{\SizeReaderTrain}{5.3}  %
\newcommand{\SizeReaderCall}{8.8 ms}  %
\newcommand{\SizePpiTotal}{49.2}  %
\newcommand{\SizePpiTrain}{27.2}  %
\newcommand{\SizePpiCall}{44–49 s}  %
\newcommand{\SizeOursCalls}{31}  %
\newcommand{\SizePpiCalls}{16}  %
\newcommand{\SizeOursMem}{0.6}  %
\newcommand{\SizePpiMem}{0.7}  %
\newcommand{\FullPpiButtons}{92.0}  %
\newcommand{\FullPpiBall}{89.3}  %
\newcommand{\FullPpiTray}{92.0}  %
\newcommand{\FullPpiBox}{96.7}  %
\newcommand{\FullPpiHandEasy}{62.7}  %
\newcommand{\FullPpiHandHard}{37.3}  %
\newcommand{\FullPpiLaptop}{46.3}  %
\newcommand{\FullPpiDrawer}{79.7}  %
\newcommand{\FullPpiDustpan}{98.7}  %
\newcommand{\FullPpiPlate}{0.0}  %
\newcommand{\FullPpiFridge}{82.6}  %
\newcommand{\FullPpiMean}{70.7}  %
\newcommand{\FullActButtons}{4.0}  %
\newcommand{\FullActBall}{38.3}  %
\newcommand{\FullActTray}{1.3}  %
\newcommand{\FullActBox}{67.0}  %
\newcommand{\FullActHandEasy}{0.0}  %
\newcommand{\FullActHandHard}{0.0}  %
\newcommand{\FullActLaptop}{0.0}  %
\newcommand{\FullActDrawer}{1.7}  %
\newcommand{\FullActDustpan}{0.0}  %
\newcommand{\FullActPlate}{0.0}  %
\newcommand{\FullActFridge}{0.0}  %
\newcommand{\FullActMean}{5.9}  %
\newcommand{\FullPeractButtons}{47.0}  %
\newcommand{\FullPeractBall}{50.0}  %
\newcommand{\FullPeractTray}{60.0}  %
\newcommand{\FullPeractBox}{62.0}  %
\newcommand{\FullPeractHandEasy}{19.7}  %
\newcommand{\FullPeractHandHard}{11.0}  %
\newcommand{\FullPeractLaptop}{36.7}  %
\newcommand{\FullPeractDrawer}{49.7}  %
\newcommand{\FullPeractDustpan}{2.0}  %
\newcommand{\FullPeractPlate}{4.0}  %
\newcommand{\FullPeractFridge}{3.0}  %
\newcommand{\FullPeractMean}{16.8}  %
\newcommand{\FullAnyBiButtons}{73.0}  %
\newcommand{\FullAnyBiBall}{36.0}  %
\newcommand{\FullAnyBiTray}{67.0}  %
\newcommand{\FullAnyBiBox}{46.0}  %
\newcommand{\FullAnyBiHandEasy}{44.0}  %
\newcommand{\FullAnyBiHandHard}{15.0}  %
\newcommand{\FullAnyBiLaptop}{7.0}  %
\newcommand{\FullAnyBiDrawer}{50.0}  %
\newcommand{\FullAnyBiDustpan}{67.0}  %
\newcommand{\FullAnyBiPlate}{8.0}  %
\newcommand{\FullAnyBiFridge}{26.0}  %
\newcommand{\FullAnyBiMean}{39.9}  %
\newcommand{\FullRotriButtonsSd}{1.73}  %
\newcommand{\FullRotriBallSd}{1.15}  %
\newcommand{\FullRotriTraySd}{1.15}  %
\newcommand{\FullRotriBoxSd}{1.73}  %
\newcommand{\FullRotriHandEasySd}{2.89}  %
\newcommand{\FullRotriHandHardSd}{2.52}  %
\newcommand{\FullRotriLaptopSd}{3.46}  %
\newcommand{\FullRotriDrawerSd}{1.73}  %
\newcommand{\FullRotriDustpanSd}{2.89}  %
\newcommand{\FullRotriPlateSd}{1.15}  %
\newcommand{\FullRotriFridgeSd}{1.73}  %
\newcommand{\FullOursAvgButtons}{89.7}  %
\newcommand{\FullOursAvgButtonsSd}{7.2}  %
\newcommand{\FullOursAvgBall}{98.0}  %
\newcommand{\FullOursAvgBallSd}{1.7}  %
\newcommand{\FullOursAvgTray}{94.3}  %
\newcommand{\FullOursAvgTraySd}{1.5}  %
\newcommand{\FullOursAvgBox}{99.0}  %
\newcommand{\FullOursAvgBoxSd}{1.0}  %
\newcommand{\FullOursAvgHandEasy}{66.0}  %
\newcommand{\FullOursAvgHandEasySd}{7.0}  %
\newcommand{\FullOursAvgHandHard}{59.3}  %
\newcommand{\FullOursAvgHandHardSd}{6.7}  %
\newcommand{\FullOursAvgLaptop}{71.3}  %
\newcommand{\FullOursAvgLaptopSd}{1.2}  %
\newcommand{\FullOursAvgDrawer}{94.7}  %
\newcommand{\FullOursAvgDrawerSd}{2.1}  %
\newcommand{\FullOursAvgDustpan}{99.0}  %
\newcommand{\FullOursAvgDustpanSd}{1.7}  %
\newcommand{\FullOursAvgPlate}{79.0}  %
\newcommand{\FullOursAvgPlateSd}{8.2}  %
\newcommand{\FullOursAvgFridge}{85.0}  %
\newcommand{\FullOursAvgFridgeSd}{6.1}  %
\newcommand{\FullOursAvgMean}{85.0}  %
\newcommand{\FullOursAvgMeanSd}{2.9}  %
\newcommand{\RangeIkAvgBox}{94.7}  %
\newcommand{\RangeIkAvgButtons}{98.0}  %
\newcommand{\RangeIkAvgTray}{98.7}  %
\newcommand{\RangeIkAvgLaptop}{91.3}  %
\newcommand{\RangeIkAvgBall}{76.7}  %
\newcommand{\RangeIkAvgHandEasy}{60.0}  %
\newcommand{\RangeIkAvgHandHard}{67.3}  %
\newcommand{\RangeIkAvgDrawer}{61.3}  %
\newcommand{\RangeIkAvgDustpan}{54.7}  %
\newcommand{\RangeIkAvgFridge}{36.7}  %
\newcommand{\RangeIkAvgPlate}{86.0}  %
\newcommand{\RangeIkAvgRope}{46.0}  %
\newcommand{\RangeIkAvgMean}{72.6}  %
\newcommand{\LooDropDrawerAvg}{39.3}  %
\newcommand{\LooDropHandHardAvg}{29.3}  %
\newcommand{\LooDropRopeAvg}{11.3}  %
\newcommand{\LooWithoutDrawerAvg}{28.0}  %
\newcommand{\LooWithoutHandHardAvg}{32.7}  %
\newcommand{\LooWithoutRopeAvg}{44.0}  %
\newcommand{\FollowPpiOwnDustpanRot}{0.95}  %
\newcommand{\FollowPpiDustpanRot}{-0.04}  %
\newcommand{\FollowOursDustpanRot}{0.88}  %
\newcommand{\FollowPpiSimCountDustpan}{0 / 20}  %
\newcommand{\FollowOursSimCountDustpan}{20 / 20}  %
\newcommand{\FollowPpiOwnLaptopRot}{1.00}  %
\newcommand{\FollowPpiLaptopRot}{0.33}  %
\newcommand{\FollowOursLaptopRot}{0.52}  %
\newcommand{\FollowPpiSimCountLaptop}{11 / 20}  %
\newcommand{\FollowOursSimCountLaptop}{20 / 20}  %
\newcommand{\FollowPpiSuccRotLaptop}{35}  %
\newcommand{\FollowOursSuccBaseLaptop}{55}  %
\newcommand{\FollowOursSuccRotLaptop}{0}  %
\newcommand{\FollowPpiBallRot}{0.06}  %
\newcommand{\FollowOursBallRot}{0.13}  %
\newcommand{\FollowPpiHandEasyRot}{-0.00}  %
\newcommand{\FollowOursHandEasyRot}{-0.00}  %
\newcommand{\FollowPpiSuccRotBox}{90}  %
\newcommand{\FollowOursSuccRotBox}{10}  %
\newcommand{\RealNoEditOneCable}{0}  %
\newcommand{\RealNoEditOneDrawer}{0}  %
\newcommand{\RealNoEditOneLego}{0}  %
\newcommand{\RealNoEditOnePlate}{4}  %
\newcommand{\RealNoEditTenCable}{8}  %
\newcommand{\RealNoEditTenDrawer}{24}  %
\newcommand{\RealNoEditTenLego}{0}  %
\newcommand{\RealNoEditTenPlate}{20}  %
\newcommand{\RealNoEditTwentyFiveCable}{44}  %
\newcommand{\RealNoEditTwentyFiveDrawer}{44}  %
\newcommand{\RealNoEditTwentyFiveLego}{16}  %
\newcommand{\RealNoEditTwentyFivePlate}{36}  %
\newcommand{\RealEditOneCable}{64}  %
\newcommand{\RealEditOneDrawer}{88}  %
\newcommand{\RealEditOneLego}{60}  %
\newcommand{\RealEditOnePlate}{72}  %
\newcommand{\RealEditTenCable}{88}  %
\newcommand{\RealEditTenDrawer}{92}  %
\newcommand{\RealEditTenLego}{80}  %
\newcommand{\RealEditTenPlate}{84}  %
\newcommand{\RealEditTwentyFiveCable}{100}  %
\newcommand{\RealEditTwentyFiveDrawer}{100}  %
\newcommand{\RealEditTwentyFiveLego}{88}  %
\newcommand{\RealEditTwentyFivePlate}{88}  %
\newcommand{\RealPpiFiftyCable}{20}  %
\newcommand{\RealPpiFiftyDrawer}{28}  %
\newcommand{\RealPpiFiftyLego}{48}  %
\newcommand{\RealPpiFiftyPlate}{8}  %
\newcommand{\DemoOneEditBox}{98.0}  %
\newcommand{\DemoOneNoEditBox}{54.0}  %
\newcommand{\DemoTenEditBox}{94.0}  %
\newcommand{\DemoTenNoEditBox}{90.0}  %
\newcommand{\DemoTwentyFiveEditBox}{98.0}  %
\newcommand{\DemoTwentyFiveNoEditBox}{94.0}  %
\newcommand{\DemoOneEditButtons}{98.0}  %
\newcommand{\DemoOneNoEditButtons}{26.0}  %
\newcommand{\DemoTenEditButtons}{100.0}  %
\newcommand{\DemoTenNoEditButtons}{58.0}  %
\newcommand{\DemoTwentyFiveEditButtons}{100.0}  %
\newcommand{\DemoTwentyFiveNoEditButtons}{100.0}  %
\newcommand{\DemoOneEditTray}{96.0}  %
\newcommand{\DemoOneNoEditTray}{24.0}  %
\newcommand{\DemoTenEditTray}{100.0}  %
\newcommand{\DemoTenNoEditTray}{92.0}  %
\newcommand{\DemoTwentyFiveEditTray}{92.0}  %
\newcommand{\DemoTwentyFiveNoEditTray}{94.0}  %
\newcommand{\DemoOneEditLaptop}{82.0}  %
\newcommand{\DemoOneNoEditLaptop}{6.0}  %
\newcommand{\DemoTenEditLaptop}{82.0}  %
\newcommand{\DemoTenNoEditLaptop}{72.0}  %
\newcommand{\DemoTwentyFiveEditLaptop}{86.0}  %
\newcommand{\DemoTwentyFiveNoEditLaptop}{88.0}  %
\newcommand{\DemoOneEditBall}{78.0}  %
\newcommand{\DemoOneNoEditBall}{50.0}  %
\newcommand{\DemoTenEditBall}{66.0}  %
\newcommand{\DemoTenNoEditBall}{86.0}  %
\newcommand{\DemoTwentyFiveEditBall}{58.0}  %
\newcommand{\DemoTwentyFiveNoEditBall}{76.0}  %
\newcommand{\DemoOneEditHandEasy}{70.0}  %
\newcommand{\DemoOneNoEditHandEasy}{0.0}  %
\newcommand{\DemoTenEditHandEasy}{50.0}  %
\newcommand{\DemoTenNoEditHandEasy}{32.0}  %
\newcommand{\DemoTwentyFiveEditHandEasy}{58.0}  %
\newcommand{\DemoTwentyFiveNoEditHandEasy}{56.0}  %
\newcommand{\DemoOneEditHandHard}{56.0}  %
\newcommand{\DemoOneNoEditHandHard}{0.0}  %
\newcommand{\DemoTenEditHandHard}{90.0}  %
\newcommand{\DemoTenNoEditHandHard}{46.0}  %
\newcommand{\DemoTwentyFiveEditHandHard}{88.0}  %
\newcommand{\DemoTwentyFiveNoEditHandHard}{66.0}  %
\newcommand{\DemoOneEditDrawer}{70.0}  %
\newcommand{\DemoOneNoEditDrawer}{6.0}  %
\newcommand{\DemoTenEditDrawer}{42.0}  %
\newcommand{\DemoTenNoEditDrawer}{14.0}  %
\newcommand{\DemoTwentyFiveEditDrawer}{74.0}  %
\newcommand{\DemoTwentyFiveNoEditDrawer}{50.0}  %
\newcommand{\DemoOneEditDustpan}{50.0}  %
\newcommand{\DemoOneNoEditDustpan}{8.0}  %
\newcommand{\DemoTenEditDustpan}{94.0}  %
\newcommand{\DemoTenNoEditDustpan}{52.0}  %
\newcommand{\DemoTwentyFiveEditDustpan}{80.0}  %
\newcommand{\DemoTwentyFiveNoEditDustpan}{82.0}  %
\newcommand{\DemoOneEditFridge}{46.0}  %
\newcommand{\DemoOneNoEditFridge}{2.0}  %
\newcommand{\DemoTenEditFridge}{52.0}  %
\newcommand{\DemoTenNoEditFridge}{40.0}  %
\newcommand{\DemoTwentyFiveEditFridge}{70.0}  %
\newcommand{\DemoTwentyFiveNoEditFridge}{52.0}  %
\newcommand{\DemoOneEditPlate}{78.0}  %
\newcommand{\DemoOneNoEditPlate}{32.0}  %
\newcommand{\DemoTenEditPlate}{80.0}  %
\newcommand{\DemoTenNoEditPlate}{48.0}  %
\newcommand{\DemoTwentyFiveEditPlate}{86.0}  %
\newcommand{\DemoTwentyFiveNoEditPlate}{72.0}  %
\newcommand{\DemoOneEditRope}{66.0}  %
\newcommand{\DemoOneNoEditRope}{6.0}  %
\newcommand{\DemoTenEditRope}{62.0}  %
\newcommand{\DemoTenNoEditRope}{28.0}  %
\newcommand{\DemoTwentyFiveEditRope}{62.0}  %
\newcommand{\DemoTwentyFiveNoEditRope}{50.0}  %
\newcommand{\DemoOneEditMean}{74.0}  %
\newcommand{\DemoOneNoEditMean}{17.8}  %
\newcommand{\DemoTenEditMean}{76.0}  %
\newcommand{\DemoTenNoEditMean}{54.8}  %
\newcommand{\DemoTwentyFiveEditMean}{79.3}  %
\newcommand{\DemoTwentyFiveNoEditMean}{73.3}  %
\newcommand{\OneStageBox}{58.0}  %
\newcommand{\OneStageButtons}{48.0}  %
\newcommand{\OneStageTray}{2.0}  %
\newcommand{\OneStageLaptop}{12.0}  %
\newcommand{\OneStageBall}{30.0}  %
\newcommand{\OneStageHandEasy}{0.0}  %
\newcommand{\OneStageHandHard}{0.0}  %
\newcommand{\OneStageDrawer}{0.0}  %
\newcommand{\OneStageDustpan}{4.0}  %
\newcommand{\OneStageFridge}{0.0}  %
\newcommand{\OneStagePlate}{8.0}  %
\newcommand{\OneStageRope}{36.0}  %
\newcommand{\OneStageMean}{16.5}  %
\newcommand{\MirrorBaseSwap}{0.0}  %
\newcommand{\RealNoEditFiftyCable}{64}  %
\newcommand{\RealNoEditFiftyDrawer}{96}  %
\newcommand{\RealNoEditFiftyLego}{36}  %
\newcommand{\RealNoEditFiftyPlate}{40}  %
\newcommand{\RealEditFiftyCable}{92}  %
\newcommand{\RealEditFiftyDrawer}{100}  %
\newcommand{\RealEditFiftyLego}{80}  %
\newcommand{\RealEditFiftyPlate}{88}  %
\newcommand{\RealPpiEditFiftyCable}{0}  %
\newcommand{\RealPpiEditFiftyDrawer}{20}  %
\newcommand{\RealPpiEditFiftyLego}{0}  %
\newcommand{\RealPpiEditFiftyPlate}{0}  %
\newcommand{\LooDropHandHardSd}{11.7}  %
\newcommand{\LooDropDrawerSd}{11.0}  %
\newcommand{\LooDropRopeSd}{12.2}  %
\newcommand{\RealDistOneLegoMed}{26.5}  %
\newcommand{\RealDistOneDrawerMed}{17.2}  %
\newcommand{\RealDistTwentyFiveCableMed}{11.7}  %
\newcommand{\RealDistTwentyFiveDrawerMed}{1.9}  %
\newcommand{\RealYawCollMin}{-153.6}  %
\newcommand{\RealYawCollMax}{-100.4}  %
\newcommand{\RealYawTestMin}{-104.3}  %
\newcommand{\RealYawTestMax}{-85.3}  %
\newcommand{\RealYawSource}{-118.0}  %
\newcommand{\RealYawOutside}{19}  %
\newcommand{\RealYawTotal}{25}  %
\newcommand{\RealNoEditOneMean}{1}  %
\newcommand{\RealNoEditTenMean}{13}  %
\newcommand{\RealNoEditTwentyFiveMean}{35}  %
\newcommand{\RealNoEditFiftyMean}{59}  %
\newcommand{\RealEditOneMean}{71}  %
\newcommand{\RealEditTenMean}{86}  %
\newcommand{\RealEditTwentyFiveMean}{94}  %
\newcommand{\RealEditFiftyMean}{90}  %
\newcommand{\FullOursAvgRope}{66.0}  %
\newcommand{\FullOursAvgRopeSd}{2.6}  %
\newcommand{\FullNoEditRope}{54.0}  %
\newcommand{\PpiRealEpochs}{100}  %
\newcommand{\RealPpiFiftyMean}{26}  %
\newcommand{\RealPpiEditFiftyMean}{5}  %

%% file: sec/appendix_arxiv.tex
\clearpage
\twocolumn[{\centering\Large\textbf{Appendix}\\\vspace{1.0em}}]
\appendix

\section{Evaluation Protocol}
\label{sec:supp:protocol}
\paragraph{Test scenes.}
Each of the 50 test scenes rotates the source scene by $\theta \in [-30^\circ,30^\circ]$ and translates it by $\Delta x,\Delta y \in [-20,20]$\,cm, spread by a maximin Latin hypercube design; when the scripted expert cannot complete a designed pose, we use a backup pose, so that the expert solves every test scene. %
Another drawer is named in 33 of the 50 drawer scenes, and an item of another color in 40 of the 50 handover-hard scenes; the new rope shapes move nodes by up to 8.3\,cm. %

\paragraph{Full data and reported baselines.}
We train on 80 of the 100 training demonstrations per task, hold out the other 20 for validation and for fitting the flow perturbations of the action generator, and test on the official 100-episode split.
The values of the other policies in \cref{tab:full} are taken from Table~II of RoTri-Diff~\cite{chen2026rotridiff}: RoTri-Diff reports three seeds on 100 rollouts from randomized initial states, AnyBimanual one multi-task checkpoint, and PPI and the other policies the best checkpoint per task, so their test episodes need not match ours.
RLBench's reset can leave left-arm links up to 13\,mm off their joints; we re-seat the arms after every reset for put item in drawer and pick plate, in every row and seed. %

\paragraph{Rows without edits and calibration.}
The rows without edits omit the flow label of the pose edit. In the one-shot setting, they also omit the three flow augmentations of the action generator (rigid scene transform, anchor subset, and temporal shift; \cref{sec:method:policy}), and keep the two perturbations that model errors of the predicted flow and of the hand pose.
In the one-shot setting, for every task and row, a few scalars are calibrated on other demonstrations of the task (the motion scale $\kappa$, a moving-anchor threshold, and two error deviations of the flow generator); neither stage trains on these demonstrations. %
Each seed retrains both generators, and all rows and seeds are evaluated on the same test scenes. %

\section{VLM Range}
\label{sec:supp:vlm}
\begin{figure*}[t]
\centering
\includegraphics[width=\linewidth]{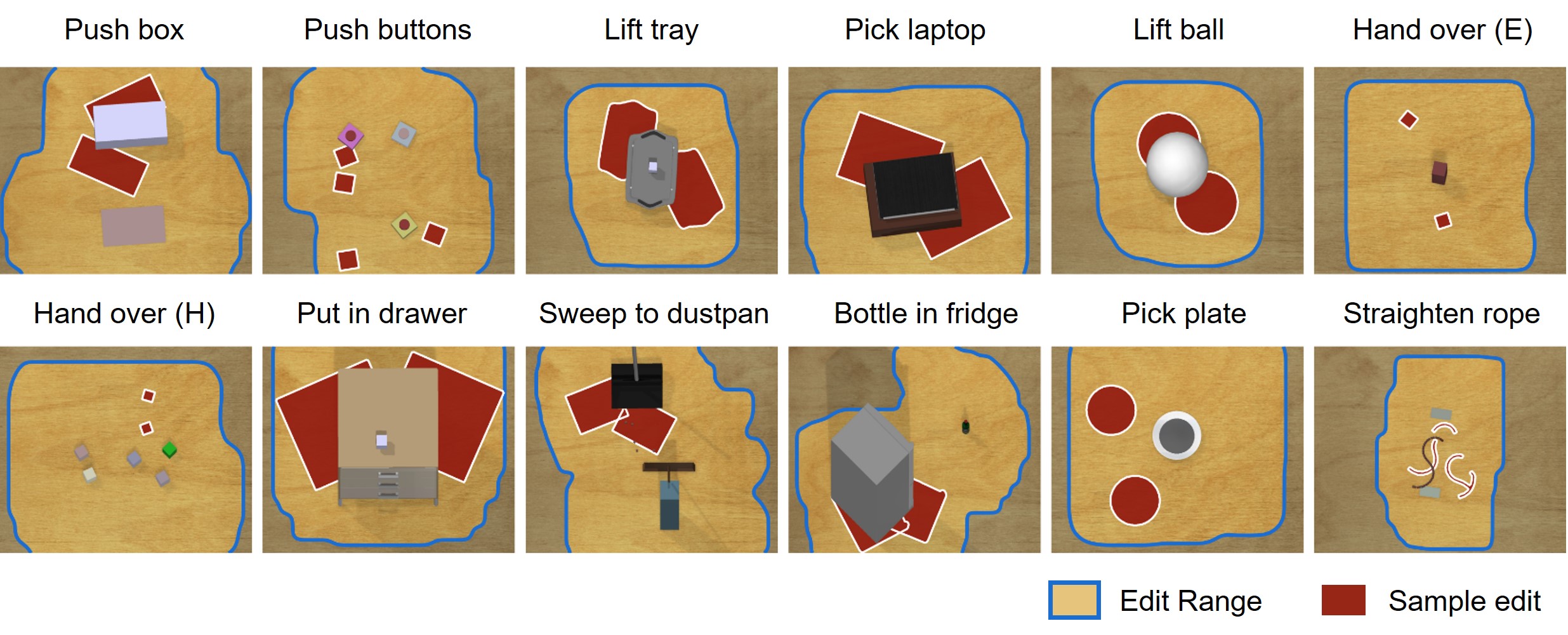}
\caption{\textbf{The VLM range of the pose edit for each task.}
Top view of the first frame of the demonstration.
Blue outline: the region that the objects occupy when the scene is moved within the VLM range.
Red regions (``Sample edit'' in the legend): two of the 50 test scenes; 44, 48, 49, and 49 of them lie inside the range for push box, push buttons, lift ball, and pick plate, and all 50 for the other tasks.} %
\label{fig:range}
\end{figure*}

The VLM receives the definition of the pose edit, a top-down rendering of the first frame with a 10\,cm grid, both arm bases, the rotation center, and the demonstrated hand positions at gripper events and pauses, a front view, and a short text on the reach of the arm, without being told which checks to use.
It writes \texttt{allowed}$(\theta,\Delta x,\Delta y)$ with the checks \texttt{reach}, \texttt{in\_view}, \texttt{on\_table}, and \texttt{moved}, which are evaluated with inverse kinematics and scene geometry, and states why each check matters for the task.
For the target names, the VLM locates each of the five color words in each of the six camera views of handover item (hard), and we keep a location seen by at least two cameras; for put item in drawer, we number the two other drawers and ask which is higher and which is the top drawer, keeping an answer only when the questions agree.

\begin{table*}[t]
\centering
\scriptsize
\setlength{\tabcolsep}{3pt}
\caption{\textbf{VLM- and IK-derived ranges give similar overall success.}
 Success rate (\%) of 1 demo + edits on the 50 test scenes of \cref{tab:oneshot}, mean over three seeds, when the pose edit draws from the range of inverse kinematics (IK) or from the VLM range. Only the action generator is trained with the other range. VLM: the 1 demo + edits row of \cref{tab:oneshot}.}
\label{tab:range}
\begin{tabular}{l*{13}{c}}
\hline
Range & \shortstack{push\\box} & \shortstack{push\\buttons} & \shortstack{lift\\tray} & \shortstack{pick\\laptop} & \shortstack{lift\\ball} & \shortstack{handover\\easy} & \shortstack{handover\\hard} & \shortstack{put in\\drawer} & \shortstack{sweep to\\dustpan} & \shortstack{bottle in\\fridge} & \shortstack{pick\\plate} & \shortstack{straighten\\rope} & Mean \\
\hline
IK & \RangeIkAvgBox & \textbf{\RangeIkAvgButtons} & \textbf{\RangeIkAvgTray} & \textbf{\RangeIkAvgLaptop} & \RangeIkAvgBall & \RangeIkAvgHandEasy & \textbf{\RangeIkAvgHandHard} & \RangeIkAvgDrawer & \textbf{\RangeIkAvgDustpan} & \RangeIkAvgFridge & \textbf{\RangeIkAvgPlate} & \RangeIkAvgRope & \RangeIkAvgMean \\
VLM (\method) & \textbf{\OneShotEditBox} & \textbf{\OneShotEditButtons} & \textbf{\OneShotEditTray} & \OneShotEditLaptop & \textbf{\OneShotEditBall} & \textbf{\OneShotEditHandEasy} & \OneShotEditHandHard & \textbf{\OneShotEditDrawer} & \OneShotEditDustpan & \textbf{\OneShotEditFridge} & \OneShotEditPlate & \textbf{\OneShotEditRope} & \textbf{\OneShotEditMean} \\
\hline
\end{tabular}
\end{table*}

\input{sec/supp_vlm_prompts_submit}

\section{Number of Demonstrations}
\label{sec:supp:demos}
\begin{table*}[t]
\centering
\scriptsize
\setlength{\tabcolsep}{3pt}
\caption{\textbf{Per-task results of \cref{fig:demos}.} Success rate (\%) on the 50 test scenes of \cref{tab:oneshot}, for one training run (seed 0), so the rows with one demonstration differ from the three-seed means of \cref{tab:oneshot}. The 10 and 25 demonstrations are nested expert demonstrations from the range of the test scenes and disjoint from them.}
\label{tab:supp:demos}
\begin{tabular}{l*{13}{c}}
\hline
Training data & \shortstack{push\\box} & \shortstack{push\\buttons} & \shortstack{lift\\tray} & \shortstack{pick\\laptop} & \shortstack{lift\\ball} & \shortstack{handover\\easy} & \shortstack{handover\\hard} & \shortstack{put in\\drawer} & \shortstack{sweep to\\dustpan} & \shortstack{bottle in\\fridge} & \shortstack{pick\\plate} & \shortstack{straighten\\rope} & Mean \\
\hline
1 demo & \DemoOneNoEditBox & \DemoOneNoEditButtons & \DemoOneNoEditTray & \DemoOneNoEditLaptop & \DemoOneNoEditBall & \DemoOneNoEditHandEasy & \DemoOneNoEditHandHard & \DemoOneNoEditDrawer & \DemoOneNoEditDustpan & \DemoOneNoEditFridge & \DemoOneNoEditPlate & \DemoOneNoEditRope & \DemoOneNoEditMean \\
10 demos & \DemoTenNoEditBox & \DemoTenNoEditButtons & \DemoTenNoEditTray & \DemoTenNoEditLaptop & \textbf{\DemoTenNoEditBall} & \DemoTenNoEditHandEasy & \DemoTenNoEditHandHard & \DemoTenNoEditDrawer & \DemoTenNoEditDustpan & \DemoTenNoEditFridge & \DemoTenNoEditPlate & \DemoTenNoEditRope & \DemoTenNoEditMean \\
25 demos & \DemoTwentyFiveNoEditBox & \textbf{\DemoTwentyFiveNoEditButtons} & \DemoTwentyFiveNoEditTray & \textbf{\DemoTwentyFiveNoEditLaptop} & \DemoTwentyFiveNoEditBall & \DemoTwentyFiveNoEditHandEasy & \DemoTwentyFiveNoEditHandHard & \DemoTwentyFiveNoEditDrawer & \DemoTwentyFiveNoEditDustpan & \DemoTwentyFiveNoEditFridge & \DemoTwentyFiveNoEditPlate & \DemoTwentyFiveNoEditRope & \DemoTwentyFiveNoEditMean \\
\hline
Single-stage, 1 demo + edits & \OneStageBox & \OneStageButtons & \OneStageTray & \OneStageLaptop & \OneStageBall & \OneStageHandEasy & \OneStageHandHard & \OneStageDrawer & \OneStageDustpan & \OneStageFridge & \OneStagePlate & \OneStageRope & \OneStageMean \\ %
1 demo + edits (\method) & \textbf{\DemoOneEditBox} & \DemoOneEditButtons & \DemoOneEditTray & \DemoOneEditLaptop & \DemoOneEditBall & \textbf{\DemoOneEditHandEasy} & \DemoOneEditHandHard & \DemoOneEditDrawer & \DemoOneEditDustpan & \DemoOneEditFridge & \DemoOneEditPlate & \textbf{\DemoOneEditRope} & \DemoOneEditMean \\
10 demos + edits & \DemoTenEditBox & \textbf{\DemoTenEditButtons} & \textbf{\DemoTenEditTray} & \DemoTenEditLaptop & \DemoTenEditBall & \DemoTenEditHandEasy & \textbf{\DemoTenEditHandHard} & \DemoTenEditDrawer & \textbf{\DemoTenEditDustpan} & \DemoTenEditFridge & \DemoTenEditPlate & \DemoTenEditRope & \DemoTenEditMean \\
25 demos + edits & \textbf{\DemoTwentyFiveEditBox} & \textbf{\DemoTwentyFiveEditButtons} & \DemoTwentyFiveEditTray & \DemoTwentyFiveEditLaptop & \DemoTwentyFiveEditBall & \DemoTwentyFiveEditHandEasy & \DemoTwentyFiveEditHandHard & \textbf{\DemoTwentyFiveEditDrawer} & \DemoTwentyFiveEditDustpan & \textbf{\DemoTwentyFiveEditFridge} & \textbf{\DemoTwentyFiveEditPlate} & \DemoTwentyFiveEditRope & \textbf{\DemoTwentyFiveEditMean} \\
\hline
\end{tabular}
\end{table*}
\Cref{tab:supp:demos} lists the per-task success rates behind \cref{fig:demos}.

\section{Training Convergence}
\label{sec:supp:conv}
\begin{figure*}[t]
\centering
\includegraphics[width=\textwidth]{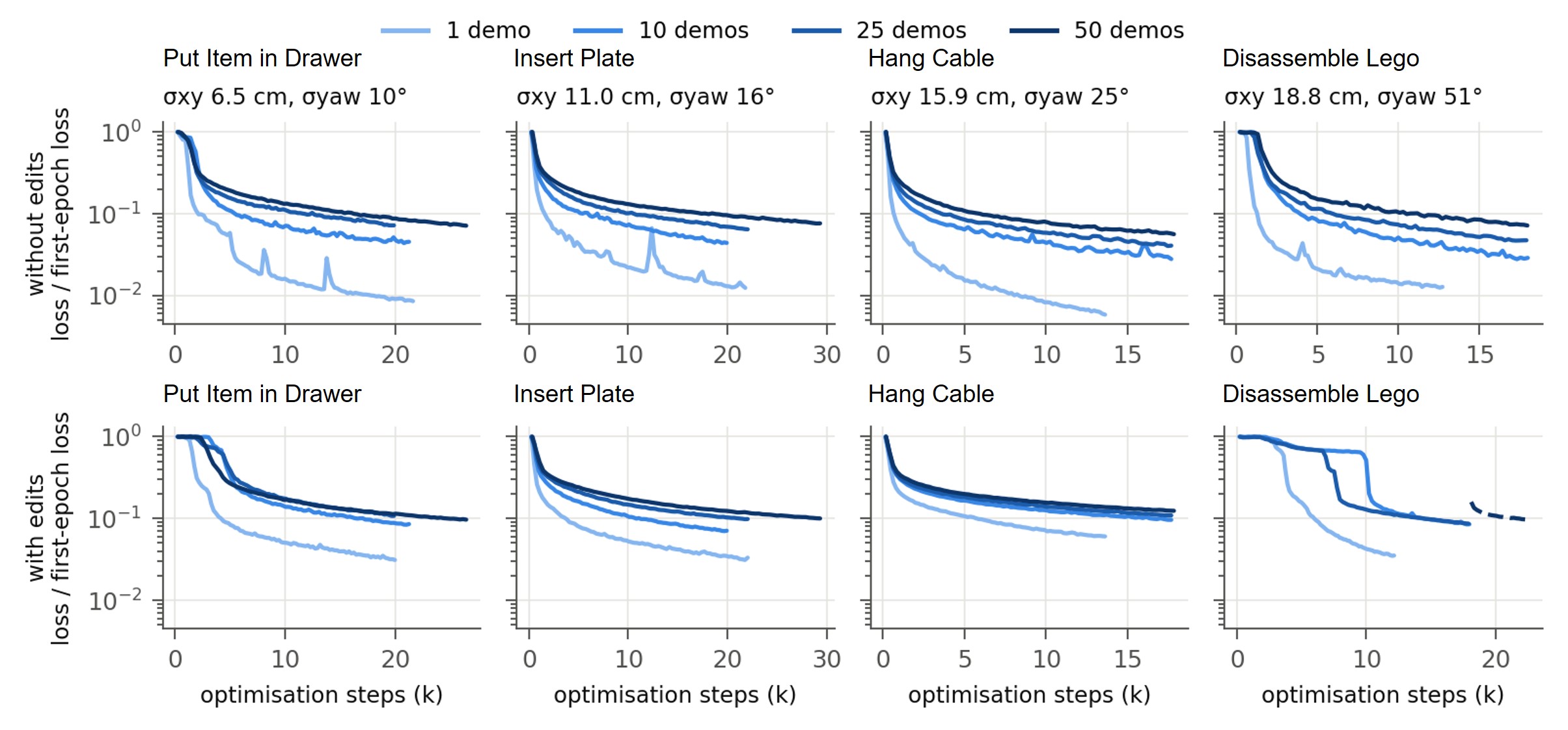}
\caption{\textbf{Edits, like more demonstrations, raise the training loss at the same step.} Training loss on the real tasks divided by that of the first epoch (log scale), without edits (top) and with edits (bottom). From left: put item in drawer, insert plate, hang cable, and disassemble Lego; $\sigma$ is the spread of the initial object placement. The dashed line (disassemble Lego, 50 demonstrations with edits) resumes from the model with 25 demonstrations and edits.}
\label{fig:supp:conv}
\end{figure*}
Edits widen the range of training scenes, as more demonstrations do, and at the same training step both raise the training loss of the flow generator on all four real tasks (\cref{fig:supp:conv}).
The two PPI rows of \cref{tab:real} train for the same \PpiRealEpochs{} epochs. With 50 demonstrations, the edits lower the success of PPI, which maps the full observation to actions with one network, on all four tasks, and raise that of \method, whose action generator learns from the flow alone (\cref{sec:exp:follow}). %

\section{Flow Interventions}
\label{sec:supp:interventions}
\begin{table}[t]
\centering
\scriptsize
\setlength{\tabcolsep}{2.4pt}
\caption{\textbf{All flow interventions.} Follow ratio $r$, mean over 20 test episodes; -- is not tested; Ours is \method. The rotation does not separate the two policies on the last two tasks. $^\ddagger$The flow predicted by PPI itself does not rotate on this task, so this cell does not test whether its action follows its flow.}
\label{tab:follow_full}
\resizebox{\columnwidth}{!}{%
\begin{tabular}{l*{8}{c}}
\hline
 & \multicolumn{2}{c}{Translation} & \multicolumn{2}{c}{Rotation} & \multicolumn{2}{c}{Instruction} & \multicolumn{2}{c}{Other ep.} \\
Task & PPI & Ours & PPI & Ours & PPI & Ours & PPI & Ours \\
\hline
put in drawer & \FollowPpiDrawerShift & \textbf{\FollowOursDrawerShift} & \FollowPpiDrawerRot$^\ddagger$ & \textbf{\FollowOursDrawerRot} & \FollowPpiDrawerInstr & \textbf{\FollowOursDrawerInstr} & \FollowPpiDrawerOther & \textbf{\FollowOursDrawerOther} \\
lift tray & \FollowPpiTrayShift & \textbf{\FollowOursTrayShift} & \FollowPpiTrayRot & \textbf{\FollowOursTrayRot} & -- & -- & \FollowPpiTrayOther & \textbf{\FollowOursTrayOther} \\
push box & \FollowPpiBoxShift & \textbf{\FollowOursBoxShift} & \FollowPpiBoxRot & \textbf{\FollowOursBoxRot} & -- & -- & \FollowPpiBoxOther & \textbf{\FollowOursBoxOther} \\
sweep to dustpan & -- & -- & $\FollowPpiDustpanRot$ & \textbf{\FollowOursDustpanRot} & -- & -- & -- & -- \\
pick laptop & -- & -- & \FollowPpiLaptopRot & \textbf{\FollowOursLaptopRot} & -- & -- & -- & -- \\
\hline
lift ball & -- & -- & \FollowPpiBallRot & \FollowOursBallRot & -- & -- & -- & -- \\
handover easy & -- & -- & $\FollowPpiHandEasyRot$ & $\FollowOursHandEasyRot$ & -- & -- & -- & -- \\
\hline
\end{tabular}}
\end{table}
Because the action head of PPI takes no flow input, we change its flow through its input: we translate or rotate the object points from which it predicts the flow, change the instruction, or use the point tokens of another episode.
For \method, we use the seed-0 model of \cref{tab:full}; for PPI, we use released checkpoints that succeed in \FollowPpiBaseDrawer, \FollowPpiBaseTray, and \FollowPpiBaseBox{} of our put item in drawer, lift tray, and push box episodes, compared with the published \FullPpiDrawer\%, \FullPpiTray\%, and \FullPpiBox\% (\cref{tab:full}).
With the rotated object points, PPI's own predicted flow follows the rotation by \FollowPpiOwnTrayRot, \FollowPpiOwnBoxRot, \FollowPpiOwnDustpanRot, and \FollowPpiOwnLaptopRot{} on lift tray, push box, sweep to dustpan, and pick laptop, while its action follows the rotated flow by \FollowPpiTrayRot, \FollowPpiBoxRot, $\FollowPpiDustpanRot$, and \FollowPpiLaptopRot{} (\cref{tab:follow}).
Translation is not diagnostic, because PPI receives the object position.
In closed loop, the hands of \method follow a rotated flow in \FollowOursSimCountDustpan{} and \FollowOursSimCountLaptop{} episodes on sweep to dustpan and pick laptop (PPI: \FollowPpiSimCountDustpan{} and \FollowPpiSimCountLaptop). %

\subsection{Single-Stage Model}
\label{sec:supp:single}
The single-stage model of \cref{sec:exp:follow} attaches the decoder of the action generator to a lighter flow generator trunk; the decoder reads the hidden state of each anchor, and the flow stays as an auxiliary output.
We train it end to end on the one-shot edits that change its observation and action labels together, which excludes moving the scene after contact and the edits that act only on the flow input of the action generator.
Its row in \cref{tab:supp:demos} is lower than \method on all 12 tasks.

\section{Distance from the Source Scene}
\label{sec:supp:ablation}
\begin{table}[t]
\centering
\scriptsize
\setlength{\tabcolsep}{4pt}
\def\pmsd#1#2{#1{\tiny$\pm$#2}}
\caption{\textbf{Flow edits keep their gain as the scene moves farther.} Success rate (\%) of the runs of \cref{tab:oneshot} over the 12 tasks, grouped by the rotation $|\theta|$ and the translation $|\Delta xy|$ of each test scene from the source scene; mean $\pm$ standard deviation over three seeds. Parentheses give the number of test scenes in each bin. Gain is in percentage points.}
\label{tab:magnitude}
\resizebox{\columnwidth}{!}{%
\begin{tabular}{lccc|lccc}
\hline
$|\theta|$ & 1 demo & + edits & Gain & $|\Delta xy|$ & 1 demo & + edits & Gain \\
\hline
$0$--$10^\circ$ (\MagThetaAN) & \pmsd{\MagThetaABase}{\MagThetaABaseSd} & \pmsd{\MagThetaAEdit}{\MagThetaAEditSd} & +\MagThetaAGain & $0$--$10$\,cm (\MagShiftAN) & \pmsd{\MagShiftABase}{\MagShiftABaseSd} & \pmsd{\MagShiftAEdit}{\MagShiftAEditSd} & +\MagShiftAGain \\
$10$--$20^\circ$ (\MagThetaBN) & \pmsd{\MagThetaBBase}{\MagThetaBBaseSd} & \pmsd{\MagThetaBEdit}{\MagThetaBEditSd} & +\MagThetaBGain & $10$--$20$\,cm (\MagShiftBN) & \pmsd{\MagShiftBBase}{\MagShiftBBaseSd} & \pmsd{\MagShiftBEdit}{\MagShiftBEditSd} & +\MagShiftBGain \\
$20$--$30^\circ$ (\MagThetaCN) & \pmsd{\MagThetaCBase}{\MagThetaCBaseSd} & \pmsd{\MagThetaCEdit}{\MagThetaCEditSd} & +\MagThetaCGain & $20$--$28$\,cm (\MagShiftCN) & \pmsd{\MagShiftCBase}{\MagShiftCBaseSd} & \pmsd{\MagShiftCEdit}{\MagShiftCEditSd} & +\MagShiftCGain \\
\hline
\end{tabular}}
\end{table}
\Cref{tab:magnitude} groups the 600 test scenes of the 12 tasks by how far they lie from the source scene. %

\section{Arm Roles}
\label{sec:supp:roles}
Since no edit swaps the two arms, \method keeps the arm roles of the demonstration.
In put item in drawer, the one-shot model thus solves \MirrorBaseSwap\% of the scenes that need the other arm to open the drawer.

\section{Real-Robot Setup and Test Scenes}
\label{sec:supp:real}
\begin{figure}[t]
\centering
\includegraphics[width=\linewidth]{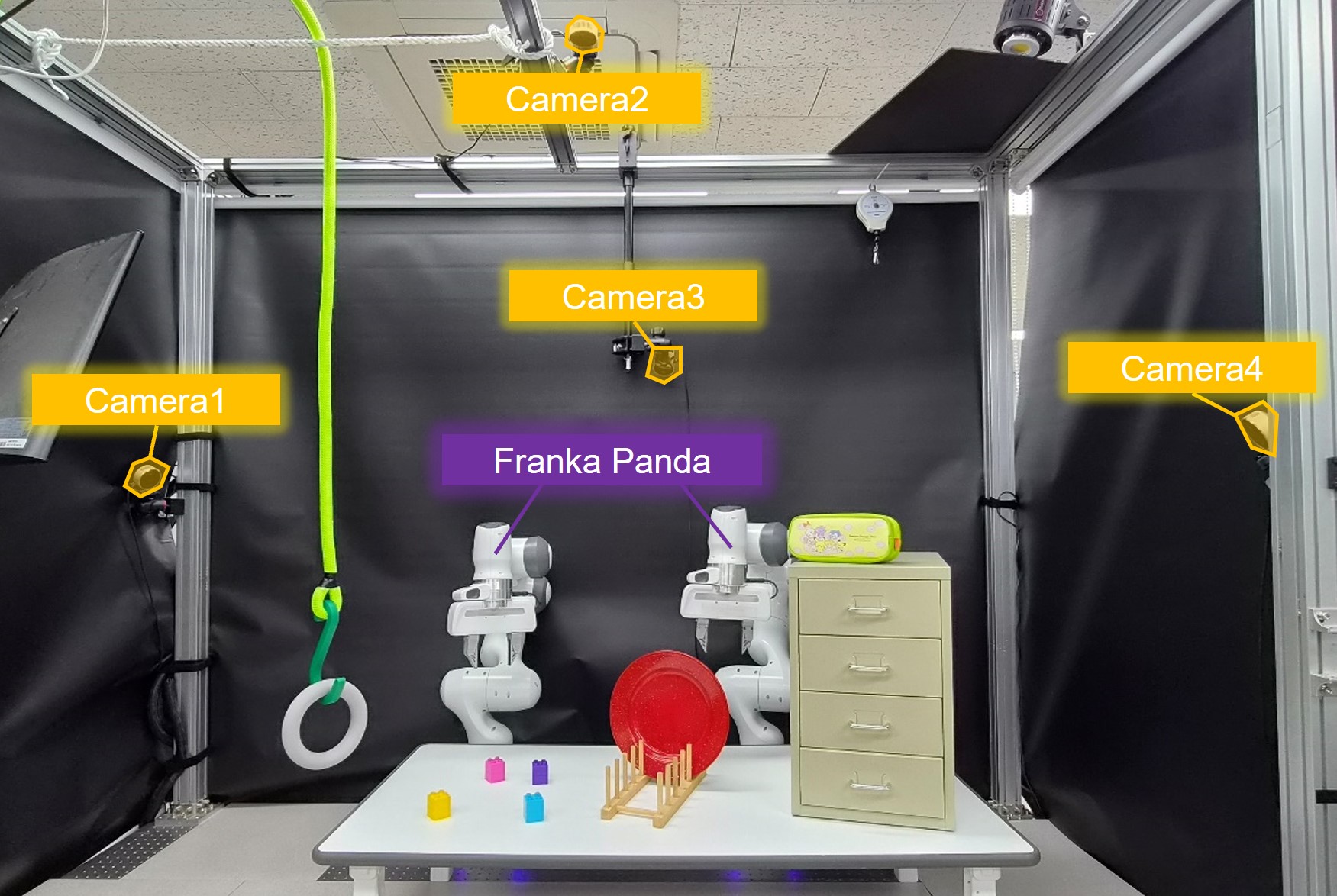}
\caption{\textbf{Real-robot setup.} Two Franka Panda arms and four RGB-D cameras. The photo shows the objects of all four tasks at once.}
\label{fig:real}
\end{figure}

\begin{figure*}[p]
\centering
\includegraphics[width=\textwidth,height=0.82\textheight,keepaspectratio]{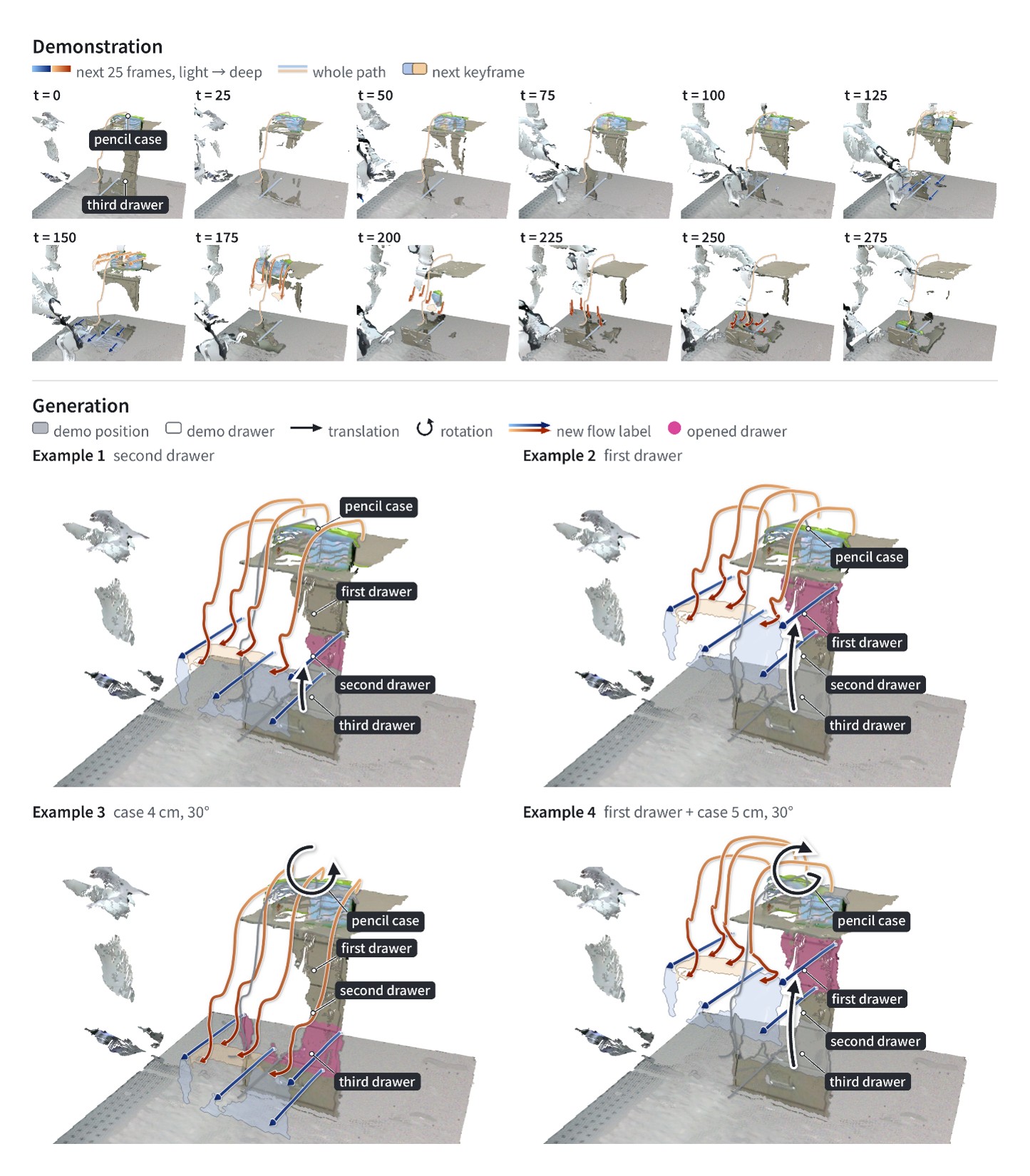}
\caption{\textbf{A real demonstration and its edits: put item in drawer.} Top: the 3D flow of the demonstration at keyframes, drawn for the next frames and for the whole path. Bottom: four edits of this demonstration. The gray shape marks where the object is in the demonstration, and the arrows show the translation and rotation of the edit. Some edits open a different drawer from the one opened in the demonstration.}
\label{fig:supp:real_drawer}
\end{figure*}

\begin{figure*}[p]
\centering
\includegraphics[width=\textwidth,height=0.82\textheight,keepaspectratio]{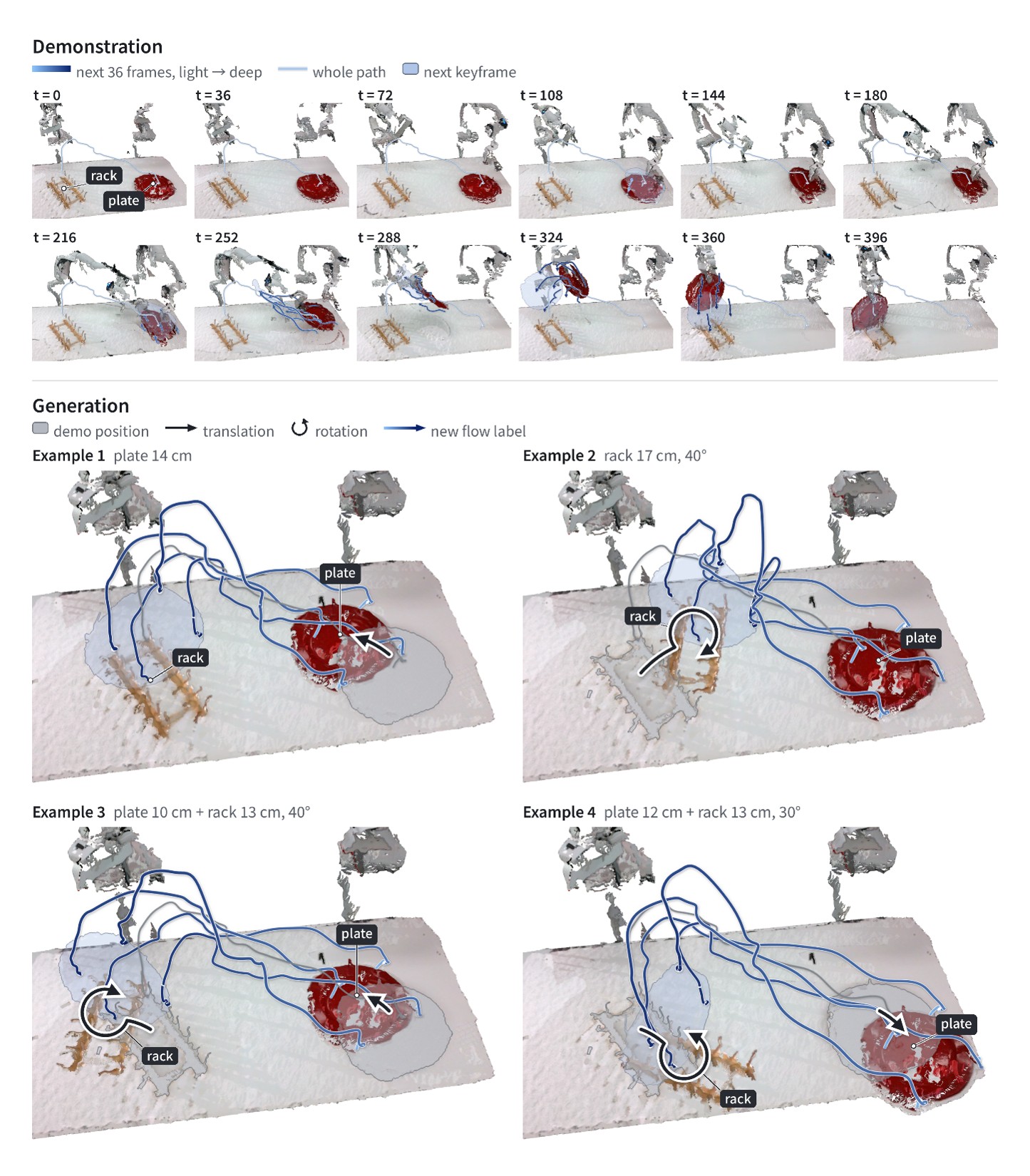}
\caption{\textbf{A real demonstration and its edits: insert plate.} Top: the 3D flow of the demonstration at keyframes, drawn for the next frames and for the whole path. Bottom: four edits of this demonstration. The gray shape marks where the object is in the demonstration, and the arrows show the translation and rotation of the edit.}
\label{fig:supp:real_plate}
\end{figure*}

\begin{figure*}[p]
\centering
\includegraphics[width=\textwidth,height=0.82\textheight,keepaspectratio]{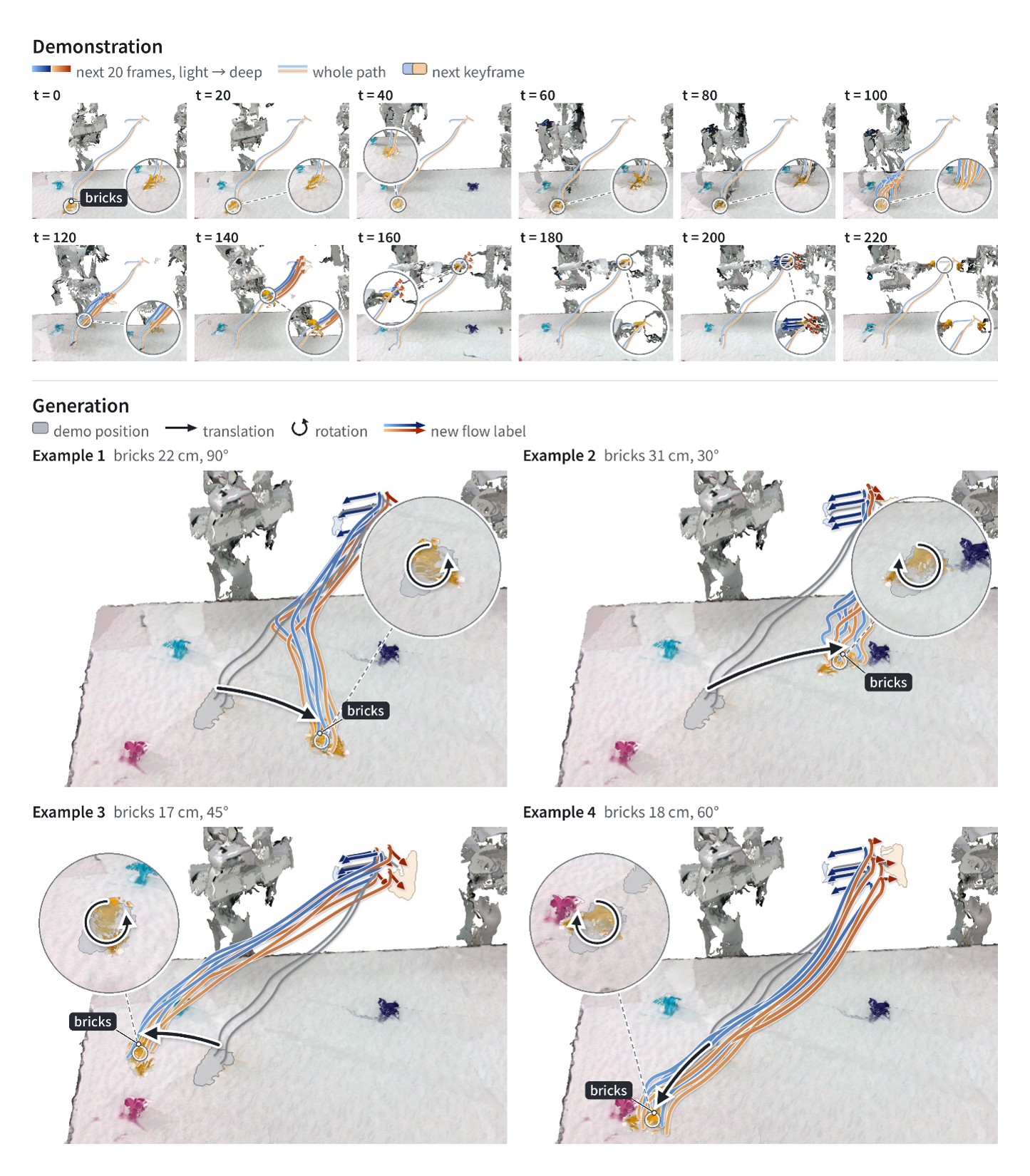}
\caption{\textbf{A real demonstration and its edits: disassemble Lego.} Top: the 3D flow of the demonstration at keyframes, drawn for the next frames and for the whole path. Bottom: four edits of this demonstration. The gray shape marks where the object is in the demonstration, and the arrows show the translation and rotation of the edit.}
\label{fig:supp:real_lego}
\end{figure*}

\begin{figure*}[p]
\centering
\includegraphics[width=\textwidth,height=0.82\textheight,keepaspectratio]{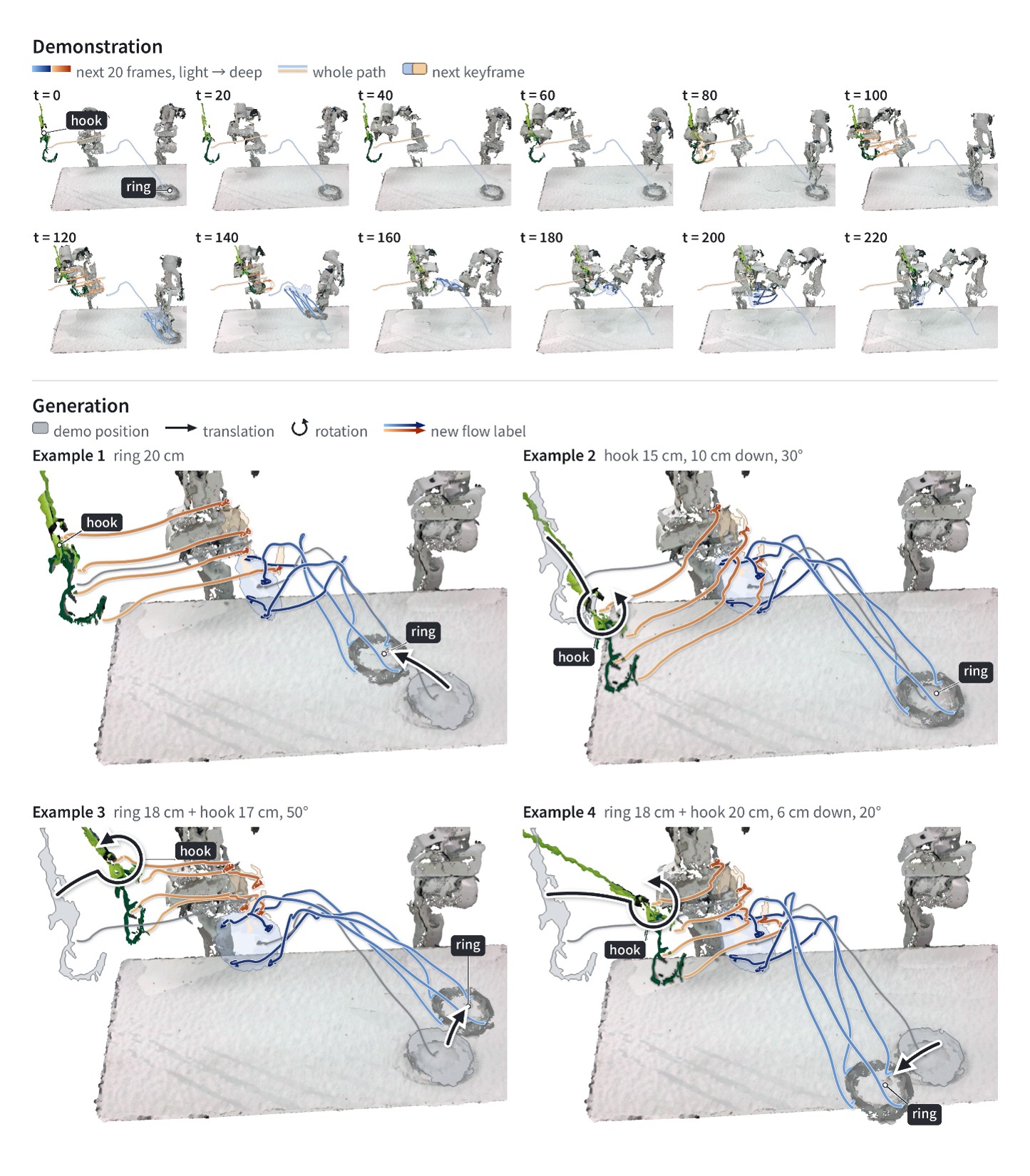}
\caption{\textbf{A real demonstration and its edits: hang cable.} Top: the 3D flow of the demonstration at keyframes, drawn for the next frames and for the whole path. Bottom: four edits of this demonstration. The gray shape marks where the object is in the demonstration, and the arrows show the translation and rotation of the edit.}
\label{fig:supp:real_cable}
\end{figure*}

We collect demonstrations with two Franka Panda arms and the bilateral force-feedback leader--follower teleoperation of FACTR~\cite{liu2025factr}, and an RTX 4080 SUPER runs the policy.
Four RGB-D cameras with extrinsic calibration observe the workspace (\cref{fig:real}), and we register their partial point clouds into one point cloud.
For the flow labels, CoTracker3~\cite{karaev2024cotracker3} tracks points in the image of each camera, and the aligned depth lifts the tracks into 3D.
\Cref{fig:supp:real_drawer,fig:supp:real_plate,fig:supp:real_lego,fig:supp:real_cable} show a demonstration of each task and four of its edits.
The pose edits of disassemble Lego and insert plate move the objects in $x$, $y$, and yaw (without yaw for the plate). In hang cable, the edit moves the hook, together with the cable end and the right hand, in $x$, $y$, $z$, and yaw within the cells where inverse kinematics finds a solution (vertically $-4$ to $+22$\,cm for the one-shot demonstration). In put item in drawer, the pose edit moves the scene in $x$ and $y$ and turns the item in yaw, and the articulation edit moves the item to another drawer of the cabinet, a vertical translation. %

\begin{figure*}[t]
\centering
\includegraphics[width=\textwidth]{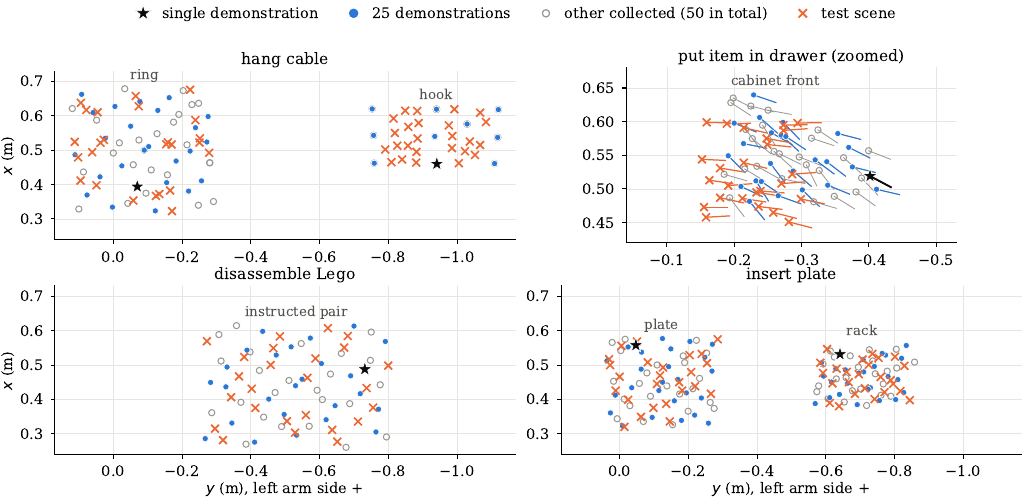}
\caption{\textbf{Planned placements of the real demonstrations and test scenes.} Top view in the robot frame ($x$ forward, $y$ toward the left arm). Each point marks the instructed Lego pair, the cabinet front, the plate and the rack, or the ring and the hook. In put item in drawer, the short line from each point shows the direction the cabinet front faces.}
\label{fig:supp:real_scenes}
\end{figure*}

Each task has 50 collected demonstrations and 25 test scenes, whose placements are planned separately from those of the demonstrations (\cref{fig:supp:real_scenes}); the training sets of 1, 10, 25, and 50 demonstrations are drawn from the collected ones.
With one demonstration, the median distance from a test scene to the nearest training scene is \RealDistOneDrawerMed--\RealDistOneLegoMed\,cm across the four tasks, and with 25 demonstrations it is \RealDistTwentyFiveDrawerMed--\RealDistTwentyFiveCableMed\,cm, measured on the planned placements. %
In put item in drawer, the test scenes turn the cabinet front to between $\RealYawTestMin^\circ$ and $\RealYawTestMax^\circ$, whereas the collected demonstrations span $\RealYawCollMin^\circ$ to $\RealYawCollMax^\circ$ and the single demonstration faces $\RealYawSource^\circ$ (planned angles about the vertical axis, $0^\circ$ along the forward axis of the robot and $90^\circ$ toward the left arm). %
The test scenes instruct each of the three drawers and each of the four Lego colors in nearly equal numbers, whereas the single demonstration of each task shows one drawer and one color.

\paragraph{Training of the real-robot rows.}
On the real robot, the contrastive term of the flow generator (\cref{sec:method:policy}) has weight 0.03 and a directional form: under the other instruction, the anchors of the instructed target should stay within 1\,cm, and those of the other target should move along the mean demonstrated direction by at least the mean demonstrated displacement; the other anchors must not change, as in simulation. It is used in every row of put item in drawer; in disassemble Lego, it is used with the color edit, in the rows with edits. In put item in drawer, where every row uses the term, the edits raise the success rate from \RealNoEditOneDrawer\% to \RealEditOneDrawer\% with one demonstration and from \RealNoEditTwentyFiveDrawer\% to \RealEditTwentyFiveDrawer\% with 25. %

\paragraph{Why the Lego rows without edits fail.}
In every Lego demonstration, the right hand turns toward the instructed pair about one second after the start, so from then on the hand position alone tells which pair moves, and only the first frames require the instruction.
Without edits and the contrastive term, the flow generator learns this shortcut: it moves no pair until the hand approaches one and then moves the pair at the hand.
We give the flow generators the logged robot inputs of the 25 test scenes with each of the four color instructions.
At the start pose, those trained on 10, 25, and 50 demonstrations without edits follow the instruction in 0, 3, and 6 of 100 inputs, and when the hand is at another pair, they move that pair in 35\%, 65\%, and 88\% of the inputs, so in this probe the shortcut grows with more demonstrations. %
With edits and the contrastive term, the flow generators follow the instruction in 79 to 91 of 100 inputs at the start pose and in all inputs with the hand at another pair. %
On the robot, the plan picks the instructed pair in 4, 13, and 9 of 25 trials without edits and in 23 to 25 trials with edits and the contrastive term. %

\section{Implementation Details}
\label{sec:supp:method_details}
\paragraph{Edits.}
The \textit{edit box} is $\theta \in [-30^\circ,30^\circ]$ and $\Delta x,\Delta y \in [-25,25]$\,cm for every task. For the flow label, 70\% of the samples move the scene by a pose drawn uniformly from the box; for the action label, 80\% of the samples draw it from the VLM range, because a moved object can leave the reach of one arm, and the other samples keep the source placement. The action label moves the scene only before the first contact, while both grippers are open, and translates without rotation for drawer and rope. Before the first contact of either hand, the anchors and their flow move while the hands stay, and each target hand trajectory blends into the moved scene within 30 steps or at first contact; in the two handover tasks, the receiving hand keeps its demonstrated targets. %
Skills apply after the point features are computed, so a moved point keeps its observed feature. The color edit exchanges the RGB values and point features of the demonstrated and target items through nearest-neighbor correspondence, applies with probability 0.8 in simulation, and changes only the flow generator's training data. The articulation edit finds $\Delta_u$ by registering the two drawers in the first frame; the item and the hand that carries it follow $c(t)\,\Delta_u$, which rises linearly from 0 at the grasp to 1 at the start of the placing approach, and the other hand reaches its full shift by the start of its own approach. %
For the deformation edit, the rope nodes are a chain fitted to the tracked rope points. A displacement $d$ comes equally often from partial straightening toward the demonstrated final shape or a smooth random bend of at most $25^\circ$ per joint; we reject displacements that move a node more than 7\,cm and pre-compute 128 of them.

\paragraph{Flow generator.}
Each scene point averages the DINOv2 features of the camera views in which it is visible; the last two DINOv2 blocks are fine-tuned with LoRA~\cite{hu2022lora}. PointNet++~\cite{qi2017pointnetpp} yields 512 scene tokens, OpenCLIP ViT-H/14~\cite{cherti2023reproducible} an instruction token, and an MLP a state token. The $M{=}200$ non-robot anchors come from farthest-point sampling, and a four-layer Transformer decoder regresses each anchor's position at every offset with an $L_1$ loss against the flow computed from the object states recorded in the demonstration and then edited.
On the two tasks whose instruction selects one of several targets, put item in drawer and handover item (hard), the flow generators of both one-shot rows add a contrastive term of weight 0.3: a second pass with the instruction of another target requires the 100-step displacements of the anchors on the two targets to differ by at least half of the mean ground-truth displacement of the instructed target, and those of all other anchors not to change. %

\paragraph{Training.}
We train the flow generator with AdamW at learning rate $10^{-4}$ ($3{\times}10^{-5}$ for LoRA, rank 8), a cosine schedule, and effective batch size 156 on two GPUs; in one-shot training, the source demonstration is repeated 80 times per epoch. The action generator uses AdamW for 48K steps at $3{\times}10^{-4}$ and batch size 64.

\paragraph{Action generator.}
In \cref{eq:anchor_token}, the factor 10 expresses displacements in decimeters, and the motion scale $\kappa=\max(1,30\,\mathrm{cm}/S)$ uses the 95th percentile $S$ of the displacements of moving anchors in the training data. The hand offsets of each anchor are embedded by a two-layer MLP whose last layer starts at zero. A four-layer Transformer encoder with 256-dimensional tokens predicts a mixture of three chunks of $K{=}50$ actions, trained with the mixture negative log-likelihood~\cite{bishop1994mixture}.
We model the flow generator's error as a common shift of the moving anchors and a progress error, matched on the held-out demonstrations (\cref{sec:supp:protocol}), and add 1\,cm per-anchor noise. The temporal shift is up to 40 steps, the anchor subset keeps 30--100\% of the anchors, the rigid scene transform is up to $\pm20^\circ$, $\pm10$\,cm horizontally, and $\pm2$\,cm vertically, and the hand-pose perturbation is 1\,cm and $4^\circ$. %

\paragraph{Model selection.}
\label{sec:supp:selection}
We select models on training demonstrations only; no test scene is used for training or model selection. %

\section{Model Size and Speed}
\label{sec:supp:cost}
\begin{table}[t]
\centering
\scriptsize
\setlength{\tabcolsep}{3pt}
\caption{\textbf{\method trains fewer parameters than PPI, and each call is over 400 times faster.} One B200 machine, batch size 1, put item in drawer, 300-step episodes.} %
\label{tab:cost}
\begin{tabular}{lcccc}
\hline
Model & \shortstack{Parameters\\(total / trained)} & \shortstack{Time\\per call} & \shortstack{Calls per\\episode} & \shortstack{Peak GPU\\memory} \\
\hline
PPI~\cite{yang2025gripper} & \SizePpiTotal\,M / \SizePpiTrain\,M & \SizePpiCall & \SizePpiCalls & \SizePpiMem\,GB \\
\method & \SizeOursTotal\,M / \textbf{\SizeOursTrain\,M} & \textbf{\SizeOursCall} & \SizeOursCalls & \textbf{\SizeOursMem\,GB} \\
\quad flow generator & \SizeGenTotal\,M / \SizeGenTrain\,M & \SizeGenCall & -- & -- \\
\quad action generator (CPU) & \SizeReaderTotal\,M / \SizeReaderTrain\,M & \SizeReaderCall & -- & -- \\
\hline
\end{tabular}
\end{table}
One call of \method takes \SizeOursCall, over 400 times faster than the \SizePpiCall{} of PPI, which runs 1000 DDPM denoising steps per call (\cref{tab:cost}).
Editing one demonstration takes \TimeScene\,ms for the flow label of the pose edit on a GPU and up to \TimeRope\,s with all edits of a task, whereas simulating one new demonstration takes about \TimeSim\,s.

%% file: sec/supp_vlm_prompts_submit.tex
\section{VLM Prompts}
\label{sec:supp:vlm_prompts}

Qwen3.5-27B~\cite{qwen2026qwen35} answers each prompt once per task, offline, with greedy decoding and thinking turned off; the policy never calls it.
The excerpts below keep the wording that the model receives, for one task where the text depends on the scene; [...] marks omitted scene details.

\paragraph{Target names.}
For handover item (hard), each camera image is asked once per color word:
\begin{Verbatim}[breaklines,breakanywhere,fontsize=\scriptsize,frame=single]
Locate "the red item" in the image. Return only a JSON list like [{"label": "red item", "bbox_2d": [x1, y1, x2, y2]}]. If it is not visible, return [].
\end{Verbatim}
For put item in drawer, the handles of the two other drawers carry the marks 1 and 2 in a crop of each camera image, and we compare the next-token scores of \texttt{1} and \texttt{2} for three questions:
\begin{Verbatim}[breaklines,breakanywhere,fontsize=\scriptsize,frame=single]
Two points in the image are marked with the numbers 1 and 2. Which mark is higher in the image, that is, closer to the top edge of the picture? Answer with the number only.
The marks 1 and 2 are on two drawers of the same cabinet. Which mark is on the top drawer? Answer with the number only.
The marks 1 and 2 are on two drawers of the same cabinet. Which mark is on the bottom drawer? Answer with the number only.
\end{Verbatim}

\paragraph{Executable range.}
The prompt holds a top-down rendering of the first frame (\cref{sec:supp:vlm}), the front camera image, and the text below, shown for lift ball:
\begin{Verbatim}[breaklines,breakanywhere,fontsize=\scriptsize,frame=single]
The edit: the whole task layout (every object of the task, together) is turned by theta degrees about the vertical axis through P (positive theta = counter-clockwise seen from above, i.e. turning +x toward +y), then shifted by dx along x and dy along y. The robots and the table do not move. After the edit the robots must do the same task, their hands following the moved objects, as in the demonstration.
Two Franka Panda arms work at a table, on opposite sides of it: [...]. Task instruction: "Lift the ball".
Image 1: the table seen from above, [...]. Image 2: the front camera.
What the arms can reach: [...] In the demonstration the hands go through these key points, which move with the layout (z = height; the table top is at z 0.75 m):
  L1: x +0.30, y +0.29, z 0.80 m (pause, step 63)
  [...]
  R5: x +0.37, y -0.14, z 1.04 m (pause, step 144)
You answer with code. These are available (theta in degrees, dx and dy in cm, as in the question):
  KEYS                            list of the key point names: L1, L2, L3, R1, R2, R3, R4, R5
  reach(name, theta, dx, dy)      True if key point `name`, moved by the edit and with its demo gripper orientation turned by theta,
                                  can be reached by its own arm (an inverse-kinematics check of the Panda arm)
  in_view(name, theta, dx, dy)    True if the moved key point is inside the image of at least 2 of the 4 fixed cameras
  on_table(name, theta, dx, dy)   True if the moved key point is above the table top
  moved(name, theta, dx, dy)      the moved key point (x, y, z) in metres
  math                            the Python math module
Write one Python function allowed(theta, dx, dy) that returns True when the two arms can still do the task with this edit.
Decide which key points and which checks matter for this task, and say why in a short reasoning first (at most 120 words).
Then give the function in one ```python code block. Use only the names above; no imports, no loops over the grid.
\end{Verbatim}
If the function uses a check other than \texttt{reach}, a second turn reports what the check tests and the share of reachable edits it removes, and asks the VLM to keep the check only if breaking it would make the task fail.
For lift ball, the VLM then drops \texttt{on\_table} and returns a function that requires every key point to be reachable.